\pdfoutput=1

\documentclass[11pt]{article}

\usepackage[final]{acl}

\usepackage{times}
\usepackage{latexsym}

\usepackage[T1]{fontenc}

\usepackage[utf8]{inputenc}

\usepackage{microtype}

\usepackage{url}
\usepackage{amsmath}
\usepackage{multirow}
\usepackage{booktabs}

\usepackage{amssymb}
\usepackage{pifont}
\usepackage{mdframed}
\usepackage{authblk}
\usepackage{inconsolata}

\usepackage{graphicx}

\newcommand{\cofirst}{$^\ast$}
\newcommand{\corresponding}{\textsuperscript{$\dagger$}}

\title{MobileVLM: A Vision-Language Model for Better Intra- and Inter-UI Understanding}

\author[1\cofirst]{\bf Qinzhuo Wu}
\author[21\cofirst]{\bf Weikai Xu}
\author[1\corresponding]{\bf Wei Liu}
\author[3]{\bf Tao Tan}
\author[1]{\bf Jianfeng Liu}
\author[1]{\\\bf Ang Li}
\author[1]{\bf Jian Luan}
\author[1]{\bf Bin Wang}
\author[2\corresponding]{\bf Shuo Shang}
\affil[1]{XiaoMi AI Lab}
\affil[2]{University of Electronic Science and Technology of China}
\affil[3]{Gaoling School of Artificial Intelligence, Renmin University of China}
\affil[ ]{\footnotesize \texttt{\{wuqinzhuo, liuwei40, liujianfeng5, liang10, luanjian, wangbin11\}@xiaomi.com}}
\affil[ ]{\footnotesize \texttt{\{xuwk266, tantao0308, jedi.shang\}@gmail.com}}

\begin{document}
\maketitle

\let\thefootnote\relax\footnotetext{\cofirst\ Equal contribution.}
\let\thefootnote\relax\footnotetext{\corresponding\ Corresponding authors.}

\begin{abstract}
Recently, mobile AI agents based on VLMs have gained increasing attention. 
These works typically utilize VLM pre-trained on general-domain data as a foundation, fine-tuning it on instruction-based mobile datasets. 
However, the proportion of mobile UI in general pre-training data is very low. Moreover, the general pre-training task does not particularly consider the characteristics of mobile UI. Therefore, directly applying such pre-trained models for mobile UI instruction fine-tuning will not yield the desired performance.
In this paper, we propose MobileVLM for Chinese UI manipulation.
On top of the general pre-training model, two additional pre-training stages are implemented with four specific tasks to enhance both intra-and inter-UI understanding.
In addition, a large Chinese mobile UI corpus,  named Mobile3M, is built from scratch to compensate for the lack of relevant data. Besides 3 million static UI pages, it also contains directed graph structures formed by real-world UI transition actions.
Experimental results show MobileVLM excels on both in-house test sets and public mobile benchmarks, outperforming existing VLMs.
Dataset and Code are available at \href{https://github.com/XiaoMi/mobilevlm}{https://github.com/XiaoMi/mobilevlm}.

\end{abstract}

\section{Introduction}

Mobile phones are widely used in daily life, and AI agents on mobile platforms are gaining industry and academic attention \cite{ding2024mobileagent, yang2023appagent}.
Due to the limitations of purely text-based LLMs in understanding User Interface (UI) elements and page structures \cite{hong2023cogagent}, recently released mobile AI agents are mainly driven by Vision-Language Models (VLM) \cite{baechler2024screenai,you2024ferret,lee2023pix2struct}. 
These works typically use VLM as a base model and then fine-tune it on instruction-based mobile datasets for domain adaptation. As a result, they excel in page navigation and can provide a coarse summary of UI functionality.

\begin{figure}[!t]
  \centering
  \includegraphics[width=0.98\columnwidth]{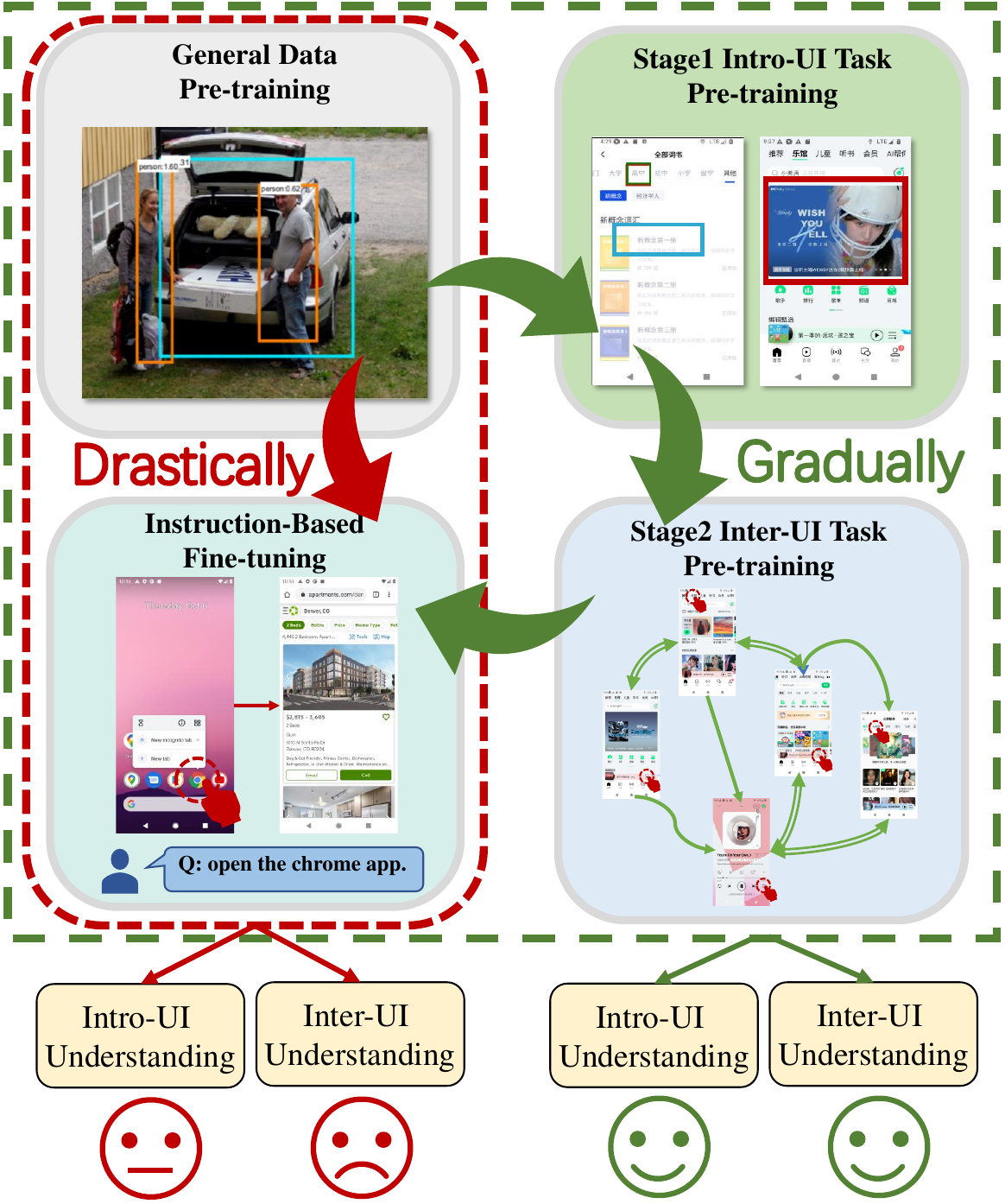}
  \caption{Previous training pipeline (red box) and ours with two additional pre-training stages (green box).}
  \label{figure1-2}
\end{figure}

\begin{table*}[!tb]
    \centering
    \resizebox{0.9\textwidth}{!}{
\begin{tabular}{lllcccll}
        \hline
        \hline
\multirow{2}{*}{Dataset} & \multirow{2}{*}{Language} &
  \multirow{2}{*}{Platform} &
  Tasks/ &
  Avg &
  Hierarchical &
  Dataset &
  \multirow{2}{*}{App Category} \\
          &    &      & Episodes & steps  & docs    & Structure &                  \\ \hline
Rico \cite{deka2017rico}    & English & Apps     & 72,219         & 1.0   & VH       & Dot       & Google \& Third-Party                \\
RicoSCA \cite{li2020mapping} & English & Apps     & 295,476         &   1.0   &  VH      & Dot         & Google \& Third-Party                \\
UIBert \cite{bai2021uibert} & English & Apps     & 16,660    & 1.0    & VH      & Dot       &Google \& Third-Party                \\
Ferret-UI \cite{you2024ferret} & English & Apps     & 320,000+       & 1.0   & VH      & Dot         &    Google \& iPhone             \\
\hline
PixelHelp \cite{li2020mapping} & English & Apps     & 187       & 4.2  & VH      & Chain         &    Google              \\
MetaGUI \cite{sun2022meta} & English & Apps     & 1,125     & 4.3  & XML     & Chain     & Google \& Third-Party                \\
Mind2Web \cite{deng2024mind2web} & English & Web      & 2,350     & 7.3   & HTML     & Chain         & n/a                \\
MoTIF \cite{burns2021mobile} & English  & Apps     & 4,707     & 4.5   & VH       & Chain         & Thrid-Party APPs \\
AITW  \cite{rawles2023android} & English   & Apps+Web & 715,142   & 6.5   & \ding{53} & Chain     & Google \& System \\
AITZ \cite{zhang2024android} & English & Apps & 2504 & 7.5  & \ding{53} & Chain     & Google \& System \\
Auto-UI \cite{zhan2023you} & English & Apps     & 687,081         & 8.3   & \ding{53}      & Chain     & Google \& System \\ \hline
Mobile3M & \textbf{Chinese} & Apps     & \textbf{3,098,786} & 6.5   & XML     & \textbf{Graph}     & Thrid-Party APPs \\ 
        \hline
        \hline
\end{tabular}
}
\caption{Comparison of Mobile3M and existing datasets. For size comparison, we list the number of apps/tasks /episodes/dialogues and average task steps. Mobile3M collects data on real and usable third-party Chinese apps.}
\label{table1}
\end{table*}

However, these VLMs like GPT-4V \cite{openai2023gpt4}, CogVLM \cite{wang2023cogvlm}, and Qwen-VL \cite{bai2023qwen} typically utilize large-scale general datasets, such as Laion-5B \cite{schuhmann2022laion}, Coyo \cite{kakaobrain2022coyo-700m}, for pre-training. 
The proportion of mobile UI pages in these datasets is very low, which results in the overall image characteristics of the datasets being quite different from those of mobile-specific datasets.

\begin{figure*}[!t]
  \centering
  \includegraphics[width=0.95\textwidth]{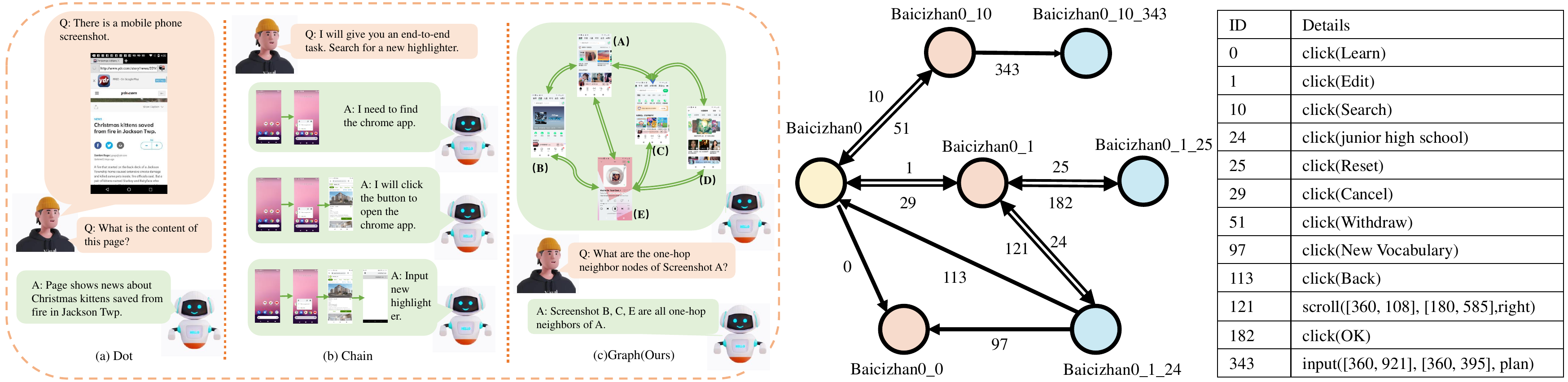}
  \caption{(Left) Dot, chain, and graph tasks from Rico, Auto-UI, and Mobile3M. (Right) A directed graph example.}
  \label{figure3-1}
\end{figure*}

Moreover, the general pre-training task does not particularly consider the characteristics of mobile UI for these VLMs. 
The general pre-training task, such as image caption and visual question answering, mainly focuses on the overall information of the image, while the mobile UI task demands capturing more fine-grained details such as layout and elements. 
As a result, these VLMs lack intra-UI information.
At the same time, these tasks only focus on the content within an image and ignore the relationship between images. 
Even for some multi-round navigation tasks in Figure \ref{figure3-1}, its interaction traces form a chain structure, which still cannot fully cover the inter-UI relationships of massive pages in a real app. Intuitively, all UI pages of an app should form a graph structure.
Therefore, these VLMs also lack inter-UI information. 

To address these issues, as shown in Figure \ref{figure1-2}, we propose two additional mobile pre-training stages and four specific mobile tasks to enhance both intra- and inter-UI understanding. 
In stage 1, 3 UI tasks are implemented to enhance the model's granular understanding of intra-UI content.
In stage 2, action prediction tasks are introduced to predict actions connecting two pages, thereby enhancing inter-UI understanding.
Based on this training framework, we propose MobileVLM, which utilizes consistent mobile data from Mobile3M for both pre-training and fine-tuning. 
This is a VLM that can simultaneously understand fine-grained element information within a UI page and the transition relationships between UI pages.



To address the lack of mobile pre-training data, we created Mobile3M, a large-scale dataset focusing on third-party Chinese apps. 
Specifically, we selected 49 popular apps and iteratively interacted with each UI element, collecting interaction traces. 
As shown in Figure \ref{figure3-1}, all interaction traces of each app are combined into a directed graph, where each node represents a UI page and each edge represents a transition action.
Eventually, Mobile3M contains millions of UI pages, XML documents, and page changes caused by user interactions.

Overall, our work has four major contributions as follows:

$\bullet$ We propose MobileVLM, the first Chinese mobile VLM, pre-trained and fine-tuned on mobile data consistently.

$\bullet$ We propose Mobile3M, the first large-scale Chinese mobile dataset with 3 million UI pages and real-world interactions, organized into a directed graph for each app.

$\bullet$ We define two extra pre-training stages and four UI-based pre-training tasks, covering both intra- and inter-UI understanding.

$\bullet$ Experimental results show that MobileVLM outperforms existing SOTA VLMs on ScreenQA (+14.34\%) and our evaluation datasets (+34.18\%).

\section{Related Work}

\subsection{Mobile UI Dataset}
Table \ref{table1} provides a comparison of multiple mobile UI datasets. At the top of the table are several ``point'' datasets. Each data instance in these datasets contains only one page, along with different fine-grained tasks and corresponding answers. 
Rico \cite{deka2017rico} is a large-scale Android UI dataset and has been widely used as a primary data source for UI modeling research. 
UIBert \cite{bai2021uibert} release two new datasets extended from Rico. 
Ferret-UI \cite{you2024ferret} uses the UI detection model \cite{zhang2021screen} to annotate fine-grained elements in Android and iPhone screens. However, these datasets only focus on the elements and layout within a single page, so it is difficult for them to capture the complete process of users using the app. 

To better reflect user behavior, several ``chained'' mobile UI data sets have been released \cite{sun2022meta,deng2024mind2web,burns2021mobile,deng2024mobile}, as shown at the bottom of Table \ref{table1}.  Each ``chain'' of data in these datasets consists of a sequence of action-UI pages. The UI page includes a screenshot and a structured document. AITW \cite{rawles2023android} is one of the largest UI control datasets with 5 subsets and 715K episodes. Auto-UI \cite{zhan2023you} further filters the GoogleApps subset in the AITW dataset, leaving 152K episodes.
However, 
these datasets only provide pages and OCR text, missing structural documents, which makes it difficult for the VLM model to learn the ability to align image and text modalities.
As shown in Figure \ref{figure3-1}, compared with chained datasets, Mobile3M's graph structure can better capture the relationship between different pages in the app.

\subsection{Mobile Vision-Language Models}
Recently, several benchmarks \cite{rawles2023android,wen2023empowering,shaw2023pixels,yan2023gpt} are proposed to evaluate page navigation and mobile phone manipulation. 
MM-Navigator\cite{yan2023gpt} and AppAgent \cite{yang2023appagent} are both GPT-4V-based agents for the page navigation task.
CogAgent\cite{hong2023cogagent} finetunes a vision-language model, CogVLM\cite{wang2023cogvlm}, to complete page navigation tasks using only screenshots as input.
UI-VLM\cite{dorka2024training} benefits from the AutoUI dataset and utilizes a sequence of past screenshots as input.

\section{Mobile3M Dataset}
In this paper, we propose Mobile3M, a large-scale dataset focusing on Chinese apps.
Mobile3M contains a total of 20,138,332 actions, covering 3,098,786 screenshots and corresponding XML documents.
These data are organized into 49 large directed graphs, each representing a mobile app, with UI pages as nodes and actions as edges.

\subsection{Background}
\begin{figure*}[!t]
  \centering
  \includegraphics[width=1\textwidth]{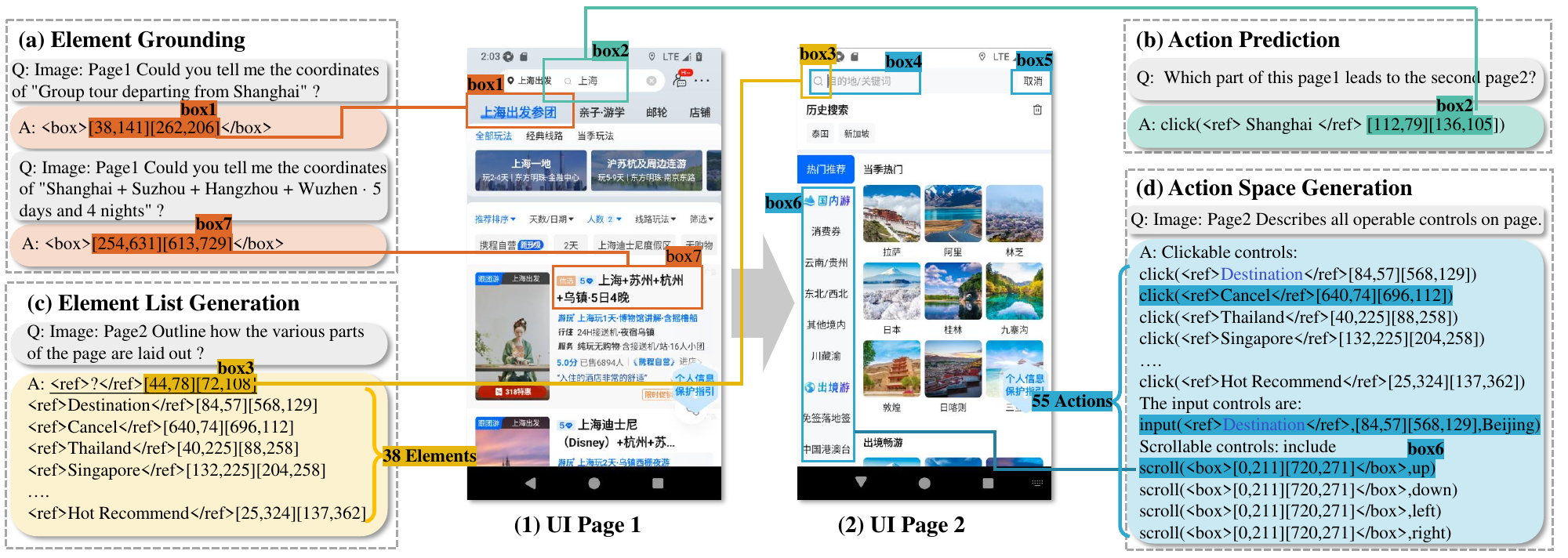}
  \caption{ Four UI-based pre-training tasks. (a)(c)(d) for stage 1 pre-training, (b) for stage 2 pre-training.}
  \label{pretrain_tasks}
\end{figure*}

\textbf{UI Page:}  We selected 49 Chinese apps from the App Store\footnote{\url{https://sj.qq.com/}}, ensuring that each app had at least 10 million downloads. 
The apps are installed and run on the emulator, and we use Appium\footnote{\url{https://appium.io/docs/en/latest/}} to collect UI pages. 
The UI page includes a screenshot and an XML document.
The XML document describes the structure and content of a UI page, including elements like buttons and text boxes, as well as layout information such as bounding boxes and hierarchical trees. 
Figure \ref{fig_UI_page} in the Appendix shows an example of a screenshot and an XML document for the app ``QQMusic''. The XML document can be parsed to produce a list of elements. As shown in the task (c) of Figure \ref{pretrain_tasks}, each element contains a name and a bounding box.

$\bullet$ Element (name, bound):  

\quad (Cancel, [640,74][696,112]).

\textbf{Action Space:} The data collection algorithm simulates the way people interact with smartphones. 
We designed three basic actions: click, scroll, and input.

$\bullet$ Click (name, bound): 

\quad click(Cancel, [640,74][696,112]).

$\bullet$ Scroll (bound, direction):  

\quad scroll([0,211][720,271],up).

$\bullet$ Input (name, bound, text): 

\quad input(Destination, [84,57][568,129], Beijing).

An element may be interactable with multiple actions. As shown in Figure \ref{pretrain_tasks}, the 38 elements in this UI page can generate an action space containing 55 actions.

\subsection{Data Collection}
Inspired by APPAgent \cite{yang2023appagent}, we use the random walk algorithm to explore apps.
The algorithm iteratively interacts with every element on each UI page and records the page transition states. 
The exploration results for each app can be represented as a directed graph, in which each edge represents an action and each node represents a UI page. 
The ``action trace'' of a UI page is defined as the shortest sequence of actions from the app's homepage to that page. 
The ID of each action in the trace is combined to create a unique identifier for the page called ``page name''.
In Figure \ref{figure3-1},  the algorithm executes ``click(Edit)'' from the ``Baicizhan0'' page to enter the Edit page.
The ID of this action is 1, 
so the name identifier of the Edit page is assigned as Baicizhan0\_1.

Since the UI pages from real-world apps may change as the app evolves.
During exploration, if the head node stored several days before is simply approximated to the current head node, UI changes may be mistaken for the edge's action, causing data errors.
Therefore, for each page, the algorithm will save screenshots and XML documents of each step in its entire ``action trace''.
Taking ``Baicizhan0\_1\_25'' in Figure \ref{figure3-1} as an example, this node contains 3 UI pages, 3 XML documents, and an ``action trace'' consisting of 2 actions.

We adopt the breadth-first method (BFS) to explore apps. Compared with the depth-first method (DFS), BFS better covers app functions and shortens the action sequence when exploring new nodes. As shown in Figure \ref{figure3-1}, the algorithm will first explore ``Baicizhan0\_10'' instead of ``Baicizhan0\_1\_25''. The task-oriented exploration method \cite{yang2023appagent} heavily relies on the performance of VLM. However, current VLMs can be costly and may perform poorly with third-party apps.
In addition, the task-oriented method may cause some infrequently used pages and app functions to be overlooked.

\subsection{Method Optimization}
The goal of building the Mobile3M dataset is to explore all functions of the app, aiming to discover new pages and actions as much as possible. 
For an app with an average action space of 50, four interactive actions will expand the app's exploration space to 6,250,000 pages, containing many duplicates.

To improve exploration efficiency, we propose a ``unique page'' mechanism. Every time a new page is explored, we use BM25 \cite{robertson2009probabilistic} to retrieve the top 5 nodes in the current app graph that are closest to the XML document of the page. The algorithm compares the new page with each of these five pages to determine if they are similar pages. The threshold of the similar coefficient is Element Diff<5 \& Pixel Diff <30\%.
Here, Element Diff is the number of different elements between two UI pages and Pixel Diff is the pixel difference between two screenshots.  
If no similar page is found in the current graph, the new page will be treated as a unique page and added to the app graph. 
As shown in Figure \ref{figure3-1}, click the ``Back'' button on ``Baicizhan0\_1\_24'', and the generated ``Baicizhan0\_1\_24\_113'' and ``Baicizhan0'' are equivalent pages. We add a directed edge from the previous page ``Baicizhan0\_1\_24'' to the similar page ``Baicizhan0'' in the graph. The algorithm will not treat the ``Baicizhan0\_1\_24\_113'' as a new explorable node.

The benefits of this mechanism are threefold:
1. This greatly reduces the exploration space of each app.  Taking ``ctrip'' as an example, our exploration process produced 187,079 UI pages with an average steps of 6.5. Without the ``unique page'' mechanism, pages of this magnitude cannot even cover all possibilities of 4-step exploration.
2. This converts the tree structure exploration results into a graph structure.  Different pages can reach ``Baicizhan0\_1'' by clicking ``Edit'', ``OK'', and scrolling. This helps the agent learn the functions of different UI elements.
3. This helps prevent the occurrence of cyclic action sequences. The ``unique page'' mechanism can detect and prune them. 

To balance the distribution of different actions in the dataset, during random walks, we give priority to the input action.
We provide 10 related keywords for each app. When executing the input action, the algorithm can randomly select a keyword to input. For scroll actions, the algorithm can choose a direction to scroll from ``up, down, left, and right''.

\begin{figure}[!t]
  \centering
  \includegraphics[width=1\columnwidth]{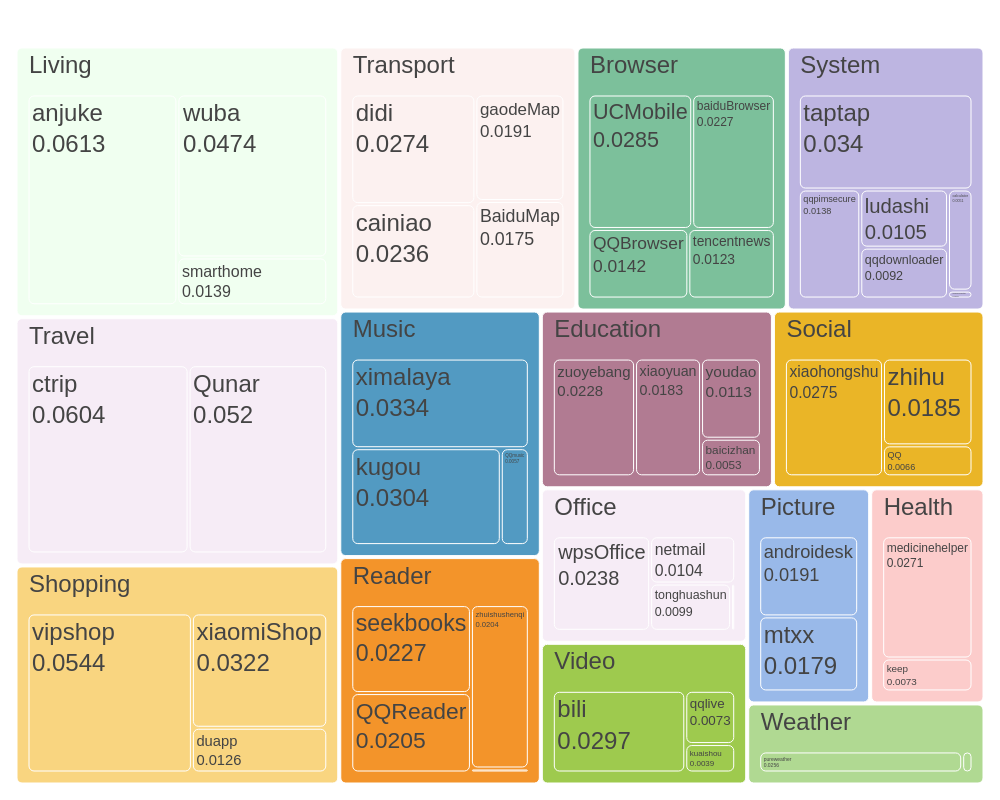}
  \caption{The proportion of data for each category and specific app in Mobile3M.}
  \label{data_percentage}
\end{figure}

\subsection{Dataset Statistics}

Among the 49 selected apps, we ensure that each main category in AppStore contains at least 2 apps.
Figure \ref{data_percentage} shows the data distribution of the Mobile3M dataset. 
The most common application categories in the dataset are ``Travel'', ``Living'' and ``Shopping''. 
As shown in the figure, Mobile3M covers multiple categories, and the amount of data in each category is relatively balanced, which ensures that the dataset is versatile and diverse.

\begin{figure}[!t]
  \centering
  \includegraphics[width=1\columnwidth]{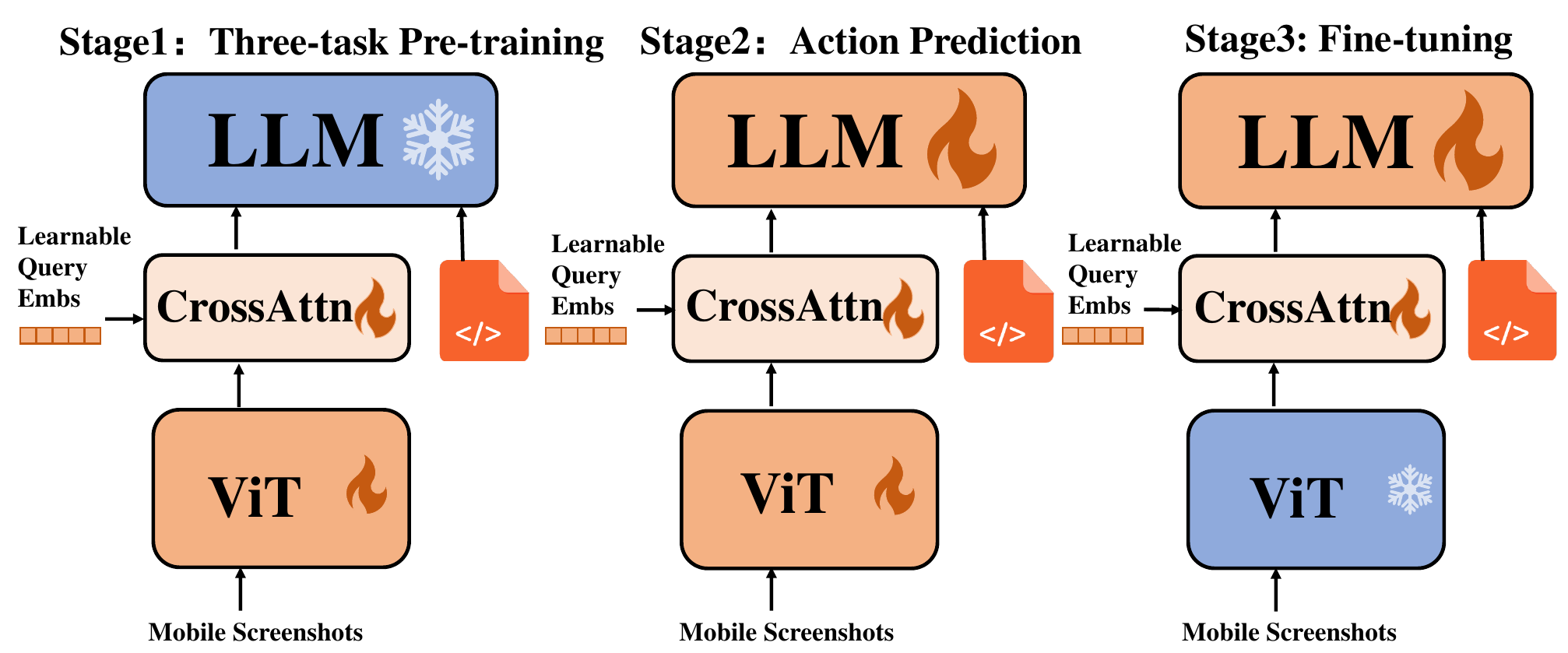}
  \caption{The three-stage pre-training and fine-tuning framework based on Qwen-VL.}
  \label{model}
\end{figure}

\section{Model}

As shown in Table \ref{tasklist}, in addition to the standard fine-tuning architecture that includes general pre-training and mobile instruction fine-tuning, we extra included two-stage mobile data pre-training and four mobile pre-training tasks.

\subsection{Pre-training}
\textbf{Stage 1: }
In the first stage of pre-training, our main goal is to enhance the VLM's understanding of intra-UI pages. We build the following three tasks to pre-train our model:

\textbf{1. Element List Generation:} This task requires models to identify all interactive elements from the page. It requires OCR and grounding abilities to recognize texts and their bounding boxes. This task provides the foundational elements for grounding and interacting in subsequent tasks.

\textbf{2. Element Grounding}~\cite{li2021vut}: The goal of this task is to enable the model to recognize and ground elements in pages. 
Given an element description, the model is required to determine its bounding box. We sample five elements on each page for grounding training.

\textbf{3. Action Space Generation:} This task requires the model to generate all candidate actions from the UI page. Based on the extracted elements, the model needs to analyze the types of elements: clickable, inputtable, and scrollable. 
This task is crucial for the action prediction tasks in Stage 2.

\textbf{Stage 2: 1. Action Prediction}
In stage 2 pre-training, we use the action prediction task to enhance VLM's ability to understand the relationship between two pages.
The expected output is the action needed to navigate from the current page to the next page.

This task aims to enhance the model's ability to predict page relationships and learn the expected outcomes of corresponding actions, providing more accurate action reasoning for downstream tasks.
In this task, the model's focus shifts from the content of intra-UIs to the complex graph structure across inter-UIs within an app.

\subsection{Fine-tuning}
\textbf{1. Page Navigation}
In Stage 3, page navigation no longer provides two pages as in Stage 2. Instead, it provides a single page along with corresponding instructions. The model needs to generate the appropriate actions based on these instructions.

\textbf{2. VQA} 
The VQA tasks require VLMs to answer the question based on a screenshot. 

In stage 3 fine-tuning, we use Mobile3M to build self page navigation task, along with Auto-UI for the page navigation task and ScreenQA for the VQA task. This stage primarily aims to convert the model's understanding of intra-UI elements and relationships between inter-UI into practical end-to-end task completion and page question-answering domain. 

\begin{table}[!tb]
    \centering
    \resizebox{1\columnwidth}{!}{
    \begin{tabular}{lc}
        \hline
        Task Name/Benchmark & Metric \\
        \hline        
		\textbf{Element List Extraction} & \\
            Self & Acc@IoU=0.1 \\
            ChineseOCRBench~\cite{liu2023hidden} & SQuAD F1* \\
            \textbf{Element Grounding} &\\
            Self & IoU=0.1 \\
            RefCOCO~\cite{veit2016coco} & IoU=0.1 \\
            \textbf{Action Space Extraction} &\\
            Self & Acc@IoU=0.1 \\
		\textbf{Action Prediction} & \\
            MoTIF-Automation~\cite{burns2021mobile} & Acc@IoU=0.1 \\
            Self & SQuAD F1*@IoU=0.1 \\
            \textbf{VQA} & \\
            ScreenQA~\cite{hsiao2022screenqa} & SQuAD F1* \\
            HumanVQA & SQuAD F1* \\
            \textbf{Page Navigation} & \\
            Auto-UI~\cite{zhan2023you} & Acc@IoU=0.1 \\ 
            Self-Navigation & Acc@IoU=0.1 \\
        \hline
    \end{tabular}
    }
    \caption{Datasets and Metrics}
    \label{task}
\end{table}

\begin{table}[!tb]
    \centering
    \resizebox{0.7\columnwidth}{!}{
    \begin{tabular}{lc}
        \hline
        \hline
        Task & \#samples \\
        \hline        
		\textbf{Mobile Pre-training: Stage 1} & \\
Element List Generation & 0.64 M\\
Element Grounding & 3.09 M\\
Action Space Generation & 0.64 M\\
\midrule
		\textbf{Mobile Pre-training: Stage 2} & \\Action Prediction & 1.92 M\\
        \midrule
		\textbf{Mobile Fine-tuning: Stage 3}   & \\
Auto-UI & 0.96 M \\
Page Navigation & 76 K\\
ScreenQA & 69 K\\
        \hline
        \hline
    \end{tabular}
    }
    \caption{Tasks in three training stages. }
    \label{tasklist}
\end{table}

\subsection{Model Architecture}
We adopt Qwen-VL-Chat~\cite{bai2023qwen-vl} as our foundation model, which consists of a Large Language Model: Qwen-7B~\cite{bai2023qwen}, a Visual Encoder: ViT-bigG~\cite{dosovitskiy2020image} with  1.9B parameters and a Position-aware Vision-Language Adapter~\cite{zhang2021tip} with 0.08B parameters.
As shown in Figure \ref{model}, we use a three-stage training method and freeze the parameters of Qwen-7B in the first stage and ViT in the third stage.

\begin{table*}[!htb]
\resizebox{2\columnwidth}{!}{
\begin{tabular}{lcccccc|cc|c|c}
\hline\hline
\multirow{2}{*}{model (stage3)} & \multicolumn{6}{c|}{Auto-UI}                 & \multicolumn{2}{c|}{Self-Navigation} & {ScreenQA} & HumanVQA \\ \cline{2-7} \cline{8-9} \cline{10-10} \cline{11-11}
                                & Overall & General & Install &  GoogleApps & Single & WebShopping & Acc  & IoU & F1 & F1     \\ \hline
BC-single  & 68.7 & - & - & - & - & - & - & - & - & -\\
BC-history  & 73.1  & 63.7 & 77.5 & 75.7 & 80.3 & 68.5 &-& -&-&- \\
\hline        
ChatGPT-CoT* & 7.72 & 5.93 & 4.38 & 10.47 & 9.39 & 8.42 & - & - & -&-\\
GPT-4V ZS+HTML* & 50.54 & 41.66 & 42.64 & 49.82 & 72.83 & 45.73 & - & - & - &- \\
GPT-4V ZS+History* & 52.96 & 43.01 & 46.14 & 49.18 & 78.29 & 48.18 & - & - & - &- \\
Qwen-VL Max & 54.15 & 46.22 & 50.30 & 49.16 & 75.32 & 49.76 & 87.2 & \underline{14.31} & \underline{71.37} & \underline{66.09} \\
GPT-4o & 55.02 & 47.06 & 49.12 & 52.30 & 80.28 & 46.42 & \underline{88.6} & 4.33 & 67.85 & 47.82\\
\hline
InternVL +History & 2.63 & 1.95 & 2.88 & 2.94 & 3.03 & 2.71 & 82.2 & 2.38  & 33.27 & 34.72\\
Qwen-VL +History & 3.23 & 2.71 & 4.11 & 4.02 & 3.89 & 2.58 & 77.4 & 4.21  & 51.51 & 52.69\\
Fine-tuned Llama 2$^{\#}$ & 28.40 & 28.56 & 35.18 & 30.99 & 27.35
& 19.92 & -& -& -& - \\
Llama 2+plan+History$^{\#}$ & 62.86 & 53.77 & 69.1 & 61.19 & 73.51 & 56.74 &- & - & - & -\\
MobileAgent & 66.92 & 55.8 & 74.98 & 63.95 & 76.27 & 63.61 & 87.4 & 7.76 & 63.76 & 47.38  \\
Auto-UI$\rm _{separate}$ & 74.07 & 65.94 & \underline{77.62} & \underline{76.45} & 81.39 & 69.72&-&-&- & - \\
Auto-UI$\rm _{unified}$ & 74.27 & 68.24 & 76.89 & 71.37 & \underline{84.58} & \underline{70.26} & - & 11.32 & - & - \\
CoCo-LLaVA  & 70.37 & 58.93 & 72.41 & 70.81 & 83.73 & 65.98 & - & - & - & - \\
CogAgent 	& 76.88 & 65.38 & 78.86	& 74.95 & \textbf{93.49}	& \underline{71.73} &  - & - & - & - \\
CoCo-Agent &  \textbf{77.82} & 	\underline{69.92}	& \textbf{80.60} & 	75.76	& \underline{88.81}	&\textbf{74.02} & - & - & - & - \\
\hline
MobileVLM w/o Stage1\&2  & 72.26 & 66.16 & 78.19 & 71.97 & 75.88 & 71.10 & 92.8 & 29.65 & 82.59 & 49.70 \\
MobileVLM w/o Stage2  & 73.05 & 70.15 & 79.41 & 74.12 & 76.26 & 41.49 & \textbf{98.2} & 35.89 & \textbf{85.71} & 76.09 \\
MobileVLM$\rm _{separate}$ & \underline{77.05} & \textbf{70.27} & 78.86 & \textbf{76.86} & 87.06 & 71.42 & - & - & 
        - & - \\
MobileVLM$\rm _{unified}$ & 74.94 & 69.58 & \underline{79.87} & 74.72 & 81.24 & 71.70 & 98 & \textbf{48.49} & 
        \textbf{85.71} & \textbf{76.82} \\
            \hline\hline
\end{tabular}
    }
    \caption{Main Result(\%). Suboptimal results are marked with an \underline{underline}. The BC results are from \citep{sun2022meta}, results with * are from \cite{ding2024mobileagent}, and results with \#\ stem from \citep{zhan2023you}. 
    Considering the testing costs, Qwen-VL-Max and GPT-4o were conducted on a random sample of 10\% of Auto-UI.}
    \label{main result}
\end{table*}

\section{Experiment}
\subsection{Datasets and Benchmarks}
We constructed our own benchmarks by selecting data from Mobile3M, and additionally selected five public Chinese benchmarks.
Specifically, we constructed the following two types of test datasets:

$\bullet$ \textbf{UnseenAPP} To verify the ability of the model on unseen apps, we selected 7 apps out of the 49 apps as shown in Table \ref{test_category} and did not use their data for training. 

$\bullet$ \textbf{SeenAPP} 
We randomly sampled 700 data for each task from the remaining 42 apps, which the model had seen during the training stage. There is no overlap between the training and the test set.


We randomly selected 500 screenshots from mobile3m and asked three annotators to construct question-and-answer pairs for each screenshot, named humanVQA benchmark.

As shown in Table \ref{task}, we choose 3 mobile benchmarks, ScreenQA and Auto-UI for evaluating stage 3 fine-tuning, and MoTIF to evaluate stage 2 pre-training.
We chose two general benchmarks, ChineseOCRBench and RefCOCO, to measure general capability loss in stage 1 pre-training.
More details can be seen in Appendix \ref{benchmarks}.

\subsection{Evaluation Metrics}
Following prior works, we used 3 objective metrics and did not use additional human evaluation.

\textbf{SQuAD F1*} For OCR and VQA tasks, we use an improved F1* score to measure the accuracy of VLM responses. 
Following OCRBench, we consider a response correct if the output contains the golden answer. Only when this condition is not met do we calculate the F1 score. F1* can be calculated as follows:
\begin{equation}
\text{F1}^* = 
\begin{cases} 
1, & \text{if Ans in GT} \\
2 \cdot \frac{\textit{Pre} \cdot \textit{Recall}}{\textit{Pre} + \textit{Recall}}, & \text{otherwise}
\end{cases}
\end{equation}

\textbf{IoU} 
Intersection over Union\cite{cheng2021boundary} is the most commonly used metric in the field of object detection.

\textbf{Action Accuracy} We follow Auto-UI's approach for evaluating action accuracy. Specifically, for click action, we allow a 14\% margin of error relative to the screen size between the predicted answers and the golden answers. For scroll action, the predicted answer only needs to be on the same axis and in the same direction as the golden answer. For input, we only calculate the F1 score of the input content.

\subsection{Implementation Details}
\textbf{Experiment Settings} 
We trained the model on NVIDIA A100 GPUs (80G×8).  
For Auto-UI finetune task, similar to its official method, we used 10\% of the GoogleApps data of AITW to save 80\% of the training time.
Our hyperparameters are as follows: learning rate of 1e-5, batch size of 4, 6000 steps for stage 1 pre-training, and 7400 steps for stage 2 pre-training.
During the testing, all baselines that were not fine-tuned were provided with few-shot instructions. More details can be seen in Appendix \ref{setting}. 
For stage 1 evaluation, we employed two SOTA models: GroundingDINO \cite{liu2023grounding} and Qwen-VL-Max. 
For Stage 2, we selected Seq2Act as the SOTA model on the MoTIF. For Stage 3, MobileVLM\(_{\text{separate}}\) were fine-tuned models based on separate subtasks of Auto-UI. MobileVLM\(_{\text{unified}}\) was the unified model for all tasks in Stage 3. For specific information on baselines, refer to Appendix \ref{baselines}.

\textbf{Data Processing} 
While Mobile3M is a Chinese dataset, Auto-UI and ScreenQA need to align with it during the testing stage. Therefore, we translated their instructions and answers into Chinese. 
Additionally, since all pages in the Mobile3M are uniformly sized at 720x1280, we resized the pages of Auto-UI and MoTIF to 720x1280. 
Our pre-training task requires VLMs to detect objects based on instructional descriptions, we removed test cases from RefCOCO that contain multiple objects in a single image to avoid ambiguity.

\begin{table*}[!htb]
    \centering
    \resizebox{1\textwidth}{!}{
\begin{tabular}{lccc|ccc|cc}
\hline
\hline
\multirow{2}{*}{model(stage1)} & Grounding & Action Space & Element List & Grounding & Action Space & Element List & RefCOCO & OCRBench(CN)\\
\cline{2-9}
          & IoU       & Acc         & Acc            & IoU       & Acc         & Acc    & IoU & Acc     \\
          \hline
InternVL  & 1.27           & 0.01          & 14.71         & 1.68   & 0.12 & 17.90        & 23.62          & 37.79         \\
Qwen-VL-Chat       & 2.92            & 0.09          & 17.32         & 2.68         & 0.04          & 19.92  & 32.37 &  35.44   \\
GroundingDINO & 16.74                 & -          & -         & 17.33             & -          & -      & \textbf{56.7} &  -  \\ \hline
Qwen-VL-Plus  & 15.25            & 1.03*          & 32.06         & 19.94  & 1.28* &  35.77        & 39.51       & 38.22         \\
Qwen-VL-Max         &  \underline{34.35}     & \underline{14.06*}          & \underline{43.79*}     &  \textbf{41.25} & 15.20* & \underline{44.91*}  & 54.61* & \textbf{47.32}      \\
GPT-4V        & 2.47         & 9.68*          & 22.49*         & 3.45        & 10.02*          & 23.02*  & - &  -       \\
GPT-4o       & 13.57       & 12.62          &   33.58      & 15.26    & \underline{16.73}       & 34.29  &  51.36* & 28.24        \\
 \hline
MobileVLM & \textbf{78.95} & \textbf{54.79} & \textbf{72.43} & 38.33 & \textbf{26.99} & \textbf{47.73} & 17.21 & 30.34 \\
\hline\hline
\end{tabular}
}
    \caption{Stage1 Result(\%). The left and middle sections show the SeenAPP and UnseenAPP from Mobile3M. The right section includes RefCOCO and ChineseOCRBench.  * indicates the test results from a 40\% random sample.} 
    \label{stage1results}
\end{table*}

\begin{table}[!htb]
    \centering
   \resizebox{0.5\textwidth}{!}{
\begin{tabular}{lcc|cc|cc}
\hline\hline
\multirow{2}{*}{model (stage2) }          & \multicolumn{4}{c|}{Action Prediction}  & \multicolumn{2}{c}{MoTIF}  \\ \cline{2-7}
& IoU     & Acc     & IoU     & Acc     & Acc     & IoU                \\ \hline
InternVL & 0.02 & 9.17 & 0.02 & 10.12 & 78.40 & 9.32\\
Qwen-VL-Chat & 0.04 & 8.2 & 0.06 & 7.34 & 81.60 & 14.22 \\
Qwen-VL Plus & 4.23 & 8.92 & 5.06 & 9.33 &- & -\\
Qwen-VL Max & \underline{10.06} & 17.32 & \textbf{12.62} & 19.69 &- & -\\
GPT-4o & 2.46 & \underline{31.23} & 3.04 &\textbf{35.07} & 93.62 & 56.40 \\
Seq2Act & - & - & - & - & \underline{99.20} & \textbf{66.40} \\
\hline
MobileVLM & \textbf{35.85} & \textbf{49.34} & 9.80 & 25.87 & \textbf{99.60} & 40.32 \\ 
\hline
\end{tabular}
}
    \caption{Stage2 Result(\%). The left part is SeenAPP. The right part is UnseenAPP. }
    \label{stage2result}
\end{table} 

\subsection{Main results}
As shown in Table \ref{main result}, MobileVLM achieved an overall improvement of 2.78\% and outperformed the Auto-UI SOTA model in all tasks. This indicates that the two-stage pre-training tasks enhanced the model's accuracy in estimating expected actions in page navigation tasks. Notably, MobileVLM achieved this despite the translation information loss and the absence of a prompt pipeline.
MobileVLM$\rm _{separate}$ slightly outperformed MobileVLM$\rm _{unified}$ due to the varying features of different tasks, which can hinder simultaneous optimization. In Self-Navigation, our model significantly outperformed GPT-4o and Qwen-VL-Max (+9.4\%, \textbf{+34.18\%}), attributed to the consistent use of mobile domain data in both pre-training and fine-tuning. 
In the ScreenQA task, MobileVLM improved by 14.34\% over Qwen-VL-Max, demonstrating superior intra-UI understanding and text extraction capabilities. Without specific fine-tuning on the HumanVQA task, MobileVLM still outperformed Qwen-VL-Max by 10.73\%, showing its excellent generalization in mobile domain VQA tasks.

\subsection{Ablation Study}
Although we surpassed the baseline in Stage 3 tasks, this could be due to inherent differences in the base models' capabilities. To validate the pre-training effect, we conducted two ablation experiments: MobileVLM w/o Stage1\&2, which is fine-tuned directly on Qwen-VL, and MobileVLM w/o Stage2, which is further fine-tuned on the Stage 1 model.
As shown in Section 4 of Table \ref{main result}, compared to MobileVLM w/o Stage1\&2, MobileVLM achieved improvements of 4.79\%, 5.2\%, \textbf{18.84\%}, and 3.12\% on Auto-UI, self-navigation, and ScreenQA, respectively.
This indicates that the two-stage pre-training improved both the model's grounding and navigation capabilities. 
Compared to MobileVLM w/o Stage2, MobileVLM improved by 4\% on Auto-UI and 12.6\% in the IoU metric for Self-Navigation (from 35.89\% to 48.49\%). This highlights the importance of the Stage 2 action prediction task in enhancing the model's navigation capability by strengthening its understanding of inter-UI relationships.
Additionally, we found that Stage 2 pre-training had little impact on VQA tasks, as these tasks rely more on the model's understanding of intra-UI elements.

\subsection{Pre-training Results} 
\noindent
\textbf{Stage1 results}
MobileVLM continues with two-stage pre-training based on Qwen-VL-Chat.
As shown in Table \ref{stage1results}, compared to Qwen-VL-Chat, MobileVLM achieved significant improvements of \textbf{76.03\%, 54.7\%}, and \textbf{55.11\%} on SeenAPP.
Moreover, compared to the best baseline Qwen-VL-Max, MobileVLM improved by \textbf{44.6\%}, \textbf{40.73\%}, and \textbf{28.64\%}. This indicates MobileVLM's superior ability to extract and ground elements. 
MobileVLM improved by 35.65\%, 26.95\%, and 27.81\% on UnseenAPP compared to Qwen-VL-Chat, and it slightly outperformed Qwen-VL-Max and GPT-4o in the element list and action space accuracy metrics, only slightly lagging behind Qwen-VL-Max in the IoU for the grounding task. 
However, due to significant differences in element distribution and layout between UnseenAPP and SeenAPP, MobileVLM, despite surpassing the best baseline, cannot fully transfer abilities learned in SeenAPP to UnseenAPP.
Since general training data was not used in Stage 2, MobileVLM is weaker on general benchmarks like RefCOCO and ChineseOCRBench compared to current SOTA models GroundingDINO and Qwen-VL-Max. 
For a detailed analysis, refer to Appendix \ref{casestudy}. 

\noindent
\textbf{Stage2 results}
As seen in the SeenAPP results in Table \ref{stage2result}, MobileVLM improved by 35.81\% and 41.14\% compared to Qwen-VL-Chat, and outperformed Qwen-VL-Max and GPT-4o by 25.79\% and 18.11\%, respectively. This indicates that the model can better understand the graph structure relationships between pages in SeenAPP.
Our model shows a certain improvement compared to Qwen-VL-Chat, but due to the significant differences in the page graph structures between UnseenAPP and SeenAPP, it is weaker than GPT-4o in recognizing the positional relationships of these apps' pages.
Nevertheless, we observed that MobileVLM exceeded Qwen-VL-Chat by 26.1\% in the IoU metric and demonstrated excellent generalization in the acc task on MoTIF (\textbf{99.6\%}).

\subsection{Language Environment Analysis.} 
In the main experiments, we translate the instructions into Chinese to align with the language we use in the pre-training stage. 
However, Qwen-vl-chat's navigation capabilities may be influenced when the environment changes back to English. 
Therefore, we supplement an experiment on the original Auto-UI dataset to compare the performance loss caused by the language environment. 
\begin{table}[h]
\centering
 \resizebox{0.5\textwidth}{!}{
\begin{tabular}{l|ccccc}
\hline
\textbf{Model} & General & Install & Single & Google apps & Web \\ 
\hline
MobileVLM\_cn & 68.87 & 79.49 & 67.38 & 74.85 & 70.85 \\
MobileVLM\_en & 69.58 & 79.87 & 81.24 & 74.72 & 71.70 \\
\hline
\end{tabular}
}
\caption{Performance Comparison of Different Languages. CN means instructions are translated into Chinese, while EN means not.}
\label{language}
\end{table}
From Table \ref{language}, it can be seen that most of the English results are very close to the Chinese results, except for some metrics. 
This shows that MobileVLM is a basic model for performing complex mobile end-to-end navigation tasks in a Chinese-English bilingual environment.

\subsection{Resource Consumption Analysis}
The inference consumption of MobileVLM under 4-bit quantization is approximately 23GB of GPU memory. Due to the need for memory to store intermediate computation results and .eval during inference, the memory requirement is generally 2-3 times that of the GPU memory. Therefore, if running MobileVLM on the device side (directly on a real phone to control apps), the model inference would require at least an 8-core CPU, 46GB of RAM, and 23GB of dedicated GPU memory. 

We designed an additional experiment: over a period of 4 minutes, with each 30-second interval as a timestamp, we recorded the memory and GPU consumption during model inference. See the table \ref{ram} below:
\begin{table}[h]
\centering
\resizebox{0.5\textwidth}{!}{
\begin{tabular}{c|p{0.2\textwidth}p{0.2\textwidth}}
\hline
\textbf{Timestamp (s)} & \textbf{RAM (MB)} & \textbf{GPU Memory (MB)} \\ 
\hline
30  & 2009.87  & 18502.0 \\
60  & 3923.33  & 20310.0 \\
90  & 3940.03  & 20798.0 \\
120 & 18438.99 & 21550.0 \\
150 & 20312.5  & 23128.0 \\
180 & 28906.25 & 23128.0 \\
210 & 39471.97 & 23128.0 \\
240 & 38438.99 & 23128.0 \\
\hline
\end{tabular}
}
\caption{Memory Usage Over Time}
\label{ram}
\end{table}

\section{Discussion}
\subsection{Can MobileVLM be applied in real applications?}
There are three main challenges to addressing the practical deployment issues of the model: 1. Resource issues, 2. Inference speed, 3. Permission issues. Below, we will discuss them in detail:
\begin{enumerate}
\item Resource Issues: As we know, current on-device large models typically range from 700M to 4.5B parameters. For example, Apple's OpenELM has 2.7B parameters, and Microsoft's Phi-3-mini has 3.8B parameters. Both models are specifically designed for mobile terminals. Our model has 9.8B parameters, which presents a significant resource requirement gap for on-device deployment. However, the smaller parameter models mentioned also have performance demand gaps.
\item Inference Speed Issues: Real-world mobile usage scenarios require quick model responses. For instance, common human-computer interaction validation requires completing clicks or swipes within a specified time. Additionally, when I need to view a scrolling advertisement, I must click it just as the target ad page scrolls from the background to the main interface. Both scenarios demand that the model completes inference within the specified time.
\item Permission Issues: As introduced in the above sections, VLM primarily relies on Appium and ADB to control the mobile device. When the model needs to be directly deployed on the phone to control apps, the dependent app, such as Siri, needs to have at least system-level permissions. This means that the app must have the current system's system signature. However, for most closed-source mobile operating systems, granting system-level signatures to third-party apps is almost impossible.
\end{enumerate}

\section{Conclusion}
We propose MobileVLM, a specialized Chinese vision-language model for mobile UI manipulation. It surpasses both open-source mobile VLMs and larger closed-source general models on multiple mobile public benchmarks. Meanwhile, we build the first large-scale Chinese mobile dataset, Mobile3M, which includes multiple pre-training and fine-tuning tasks specific to mobile scenarios. We hope this work will promote the development of vision-language models in the mobile domain and provide a reference for future Mobile-agent research.

\clearpage
\newpage

\section*{Limitations}
Our training data includes 49 commonly used apps, but this may still not fully cover all scenarios of daily life, due to the vastness of the Android app market. In future work, we will continue to expand the number of apps. Additionally, because some apps have extra paid content, such as VIP, our model may not have fully learned all their functionalities. Our data may also have some temporal limitations, as random app updates can cause changes in page and action traces.

\section*{Ethics Statement}
Our training data does indeed contain some personal information of the authors, but we commit to anonymizing all private data before making it public. Additionally, the personal information in the data before anonymization has been authorized by the respective individuals for use during the training stage.
In the process of generating manually annotated data through crowdsourcing, we employed seven employees from a crowdsourcing company without discrimination. During the annotation process, they were provided with corresponding mobile screenshots and structured texts, and we paid them labor compensation of no less than 120 CNY per hour.

\section*{Acknowledgements}
We thank the Xiaomi SmartPhone Department of Xiaomi Technology Corporation for their emulator support for this project.
This work was supported by the NSFC (U2001212, 62032001, and 61932004).



\bibliography{acl_latex}

\clearpage
\newpage
\onecolumn 
\appendix
\section{Experiment Settings}{\label{setting}}
\subsection{Baselines}\label{baselines}
Our baseline models are as follows: 
\begin{itemize}
    \item Specialized UI Agent. We adopted the Behavioral Cloning (BC) agent. BC is a Transformer-based architecture that takes a task instruction, the current screen, and a stacked history of screen observations and actions as input. The task instruction and OCR-detected texts are encoded by a pre-trained BERT. The icons are represented by the embeddings for each of the bounding box points. The screen history is modeled by the {x, y} positions of the touch and lift actions. All the embedded representations are fused to predict the action by a decoder. There are two BC variants, BC-single and BC-history, depending on whether the model takes as input the screen-action history. 
    \item Fine-tuned LLMs. We follow Auto-UI, adopt Llama 2 as the baseline, and fine-tune it with LoRA. We feed the model with the user instruction and the screen descriptions in HTML syntax (the same as adopted for in-context learning LLMs). The model is expected to predict the action in the same output format as in-context learning LLMs. As fine-tuning an LLM is expensive, we randomly sample 1\% training data to help the LLM adapt to our tasks. GroundingDINO was pre-trained and fine-tuned specifically on the COCO grounding datasets~\cite{chen2024lcvo,shao2019objects365,kuznetsova2020open,xu2024multi}. Seq2Act~\cite{li2020mapping} is the state-of-the-art (SOTA) model on the MOTIF dataset.
    \item In-context Learning VLMs. We used Qwen-VL Max, Qwen-VL Plus, GPT-4o, and GPT-4v as closed-source models and provided them with few-shot examples. For specific information, refer to the next subection. Qwen-VL-Max was pre-trained specifically for Chinese OCR tasks. 
\end{itemize}
\subsection{Benchmarks}\label{benchmarks}
ScreenQA and MoTIF are VQA and no-instruction multi-step navigation benchmarks from Rico while Auto-UI is a dataset of multi-step tasks with instructions from AITW.
Since we did not mix other general data during the multi-stage pre-training, we chose two general benchmarks to measure capabilities loss.
ChineseOCRBench\footnote{\url{https://huggingface.co/datasets/SWHL/ChineseOCRBench}} is a Chinese subset of OCRBench~\cite{liu2023hidden} which consists of ESTVQA(Ch) and ReCTS(ch). RefCOCO~\citep{veit2016coco} is the most widely used object detection dataset in computer visual domain.
\subsection{Few-shot Prompt}
For the \textbf{Grounding Task}, we use this few-shot prompt to guide the VLMs in answering the questions:
\begin{mdframed}[linewidth=1pt,linecolor=black]
\textit{Here are three examples and a question. You need to help me find the location of the text I'm looking for in the image and output its bounding box. Please note that the coordinates are relative to the top-left corner of the image. Here are the three examples:}
\begin{enumerate}
    \item \textit{Question: In the image <img>\{image1\}</img>}, where is ``city'' located? \\
    Answer: <ref>city</ref><box>(24,391),(136,432)</box>
    \item \textit{Question: In the image \texttt{<img>\{image2\}</img>}, where is ``singsing'' located? \\
    Answer: <ref>singing</ref><box>(494,187),(546,223)</box>}
    \item \textit{Question: In the image \texttt{<img>\{image3\}</img>}, where is ``puss words'' located? \\
    Answer: <ref>puss words</ref><box>(483,274),(637,329)</box>}
\end{enumerate}
\textit{Now I will formally ask the question. Please note that you only need to provide the bounding box coordinates for the corresponding text.}
\end{mdframed}

For the \textbf{Navigation Task}, we use this few-shot prompt to guide the VLMs in answering the questions: 

\begin{mdframed}[linewidth=1pt,linecolor=black]
\textit{Here are two examples and a question. You need to tell me which control to interact with to navigate from the first image to the second image. There are three actions: click, input, and scroll. Here are the two examples:}
\begin{enumerate}
\item \textit{Question: Image one: <img>/home/corpus/test
\_515/few\_shot/navigation/QQmusic0\_29\_29/
QQmusic0\_29-screen.png</img>, Image two: <img>/home/corpus/test\_515/few\_shot/navig
ation/QQmusic0\_29\_29/QQmusic0\_29\_29-screen.png</img>, how to navigate from the first image to the second image?} \\
\textit{Answer:} click(<ref>GOPRO</ref><box>[200
,1132][240,1160]</box>)
\item \textit{Question: Image one: <img>/home/corpus/test
\_515/few\_shot/navigation/ctrip0\_1\_36\_313\_36
17\_1988\_3587\_3018\_13545/ctrip0\_1\_36\_313\_36
17\_1988\_3587\_3018-screen.png</img>, Image two: <img>/home/corpus/test\_515/few\_shot/navi
gation/ctrip0\_1\_36\_313\_3617\_1988\_3587
\_3018\_13545/ctrip0\_1\_36\_313\_3617\_1988
\_3587\_3018\_13545-screen.png</img>, how to navigate from the first image to the second image?} \\
\textit{Answer:} click(<ref>Thursday</ref><box>[563,163][603,185]</box>)
\end{enumerate}
\textit{Now I will formally ask the question. Please note that you need to strictly follow the example format, and you can only perform one action.}
\end{mdframed}

For the \textbf{OCR Task}, we use this few-shot prompt to guide the LLMs in answering the questions: 
\begin{mdframed}[linewidth=1pt,linecolor=black]
\begin{enumerate}
\item \textit{Question: Image: <img>/home/corpus/test\_515/few\_shot/ocr/ctrip0\_1\_36\_313\_3617\_1988\_8}
\textit{364\_4440\_1566/ctrip0\_1\_36\_313\_3617\_1988\_836} \\
\textit{4\_4440\_1566-screen.png</img>}
\textit{Explain how the elements on the current page support its content and functionality.} \\
\textit{Answer: <ref>Calendar</ref><box>[152,144][244,185]</box>} \\
\textit{<ref>Price Trend</ref><box>[421,144][575,185]</box>} \\
\textit{<ref>Calendar</ref><box>[154,205][242,211]</box>} \\
\textit{<ref>Direct Flights Only OFF</ref><box>[474,228][696,282]</box>} \\
\textit{<ref>May 2024</ref><box>[32,885][222,939]</box>} \\
\textit{<ref>April 2024</ref><box>[32,357][222,411]</box>} \\
\textit{<ref>The selected date is the departure date. The price shown is for a single adult, and frequent price changes are subject to actual payment price.</ref><box>[0,1122][720,1184]</box>}

\item \textit{Question: Image: <img>/home/corpus/test\_515/few\_shot/ocr/baicizhan0\_1\_24\_113\_159\_165} \\
\textit{\_156\_206/baicizhan0\_1\_24\_113\_159\_165\_} \\
\textit{156\_206-screen.png</img>}
\textit{Summarize the layout and interaction methods of each part of the page.} \\
\textit{Answer: <ref>All Word Books</ref><box>[0,66][720,115]</box>} \\
\textit{<ref>Popular</ref><box>[15,132][123,236]</box>} \\
\textit{<ref>University</ref><box>[123,132][231,236]</box>} \\
\textit{<ref>High School</ref><box>[231,132][339,236]</box>} \\
\textit{<ref>Middle School</ref><box>[339,132][447,236]</box>} \\
\textit{<ref>Primary School</ref><box>[447,132][555,236]</box>} \\
\textit{<ref>Study Abroad</ref><box>[555,132][663,236]</box>} \\
\textit{<ref>Others</ref><box>[663,132][720,236]</box>} \\
\textit{<ref>Popular</ref><box>[38,266][114,322]</box>} \\
\textit{<ref>College Entrance Exam Vocabulary</ref><box>[214,374][377,419]</box>} \\
\textit{<ref>Fully includes the basic and high-scoring vocabulary for the college entrance exam, suitable for students nationwide</ref><box>[214,434][682,514]</box>} \\
\textit{<ref>Total 4135 Words</ref><box>[214,524][324,558]</box>} \\
\end{enumerate}
\end{mdframed}
\begin{mdframed}[linewidth=1pt,linecolor=black]
\textit{<ref>Added</ref><box>[613,524][682,558]</box>} \\
\textit{<ref>Middle School Exam Vocabulary</ref><box>[214,604][377,649]</box>} \\
\textit{<ref>Fully includes the must-know, frequently tested, and difficult vocabulary for the middle school exam, suitable for students nationwide</ref><box>[214,664][682,744]</box>} \\
\textit{<ref>Total 2124 Words</ref><box>[214,754][324,788]</box>} \\
\textit{<ref>Added</ref><box>[613,754][682,788]</box>} \\
\textit{<ref>Complete Vocabulary for CET-4</ref><box>[214,834][439,879]</box>} \\
\textit{<ref>Fully includes the latest vocabulary for CET-4, suitable for all students preparing for the exam</ref><box>[214,894][682,974]</box>} \\
\textit{<ref>Total 4440 Words</ref><box>[214,984][324,1018]</box>} \\
\textit{<ref>Added</ref><box>[613,984][682,1018]</box>} \\
\textit{<ref>High Frequency Words for CET-4</ref><box>[214,1064][377,1109]</box>} \\
\textit{<ref>Selected high-frequency words from CET-4 real exams, helping you quickly conquer CET-4</ref><box>[214,1124][682,1184]</box>}
\end{mdframed}

\begin{figure}[!t]
  \centering
  \begin{minipage}[b]{0.40\textwidth}
    \centering
    \includegraphics[width=\textwidth]{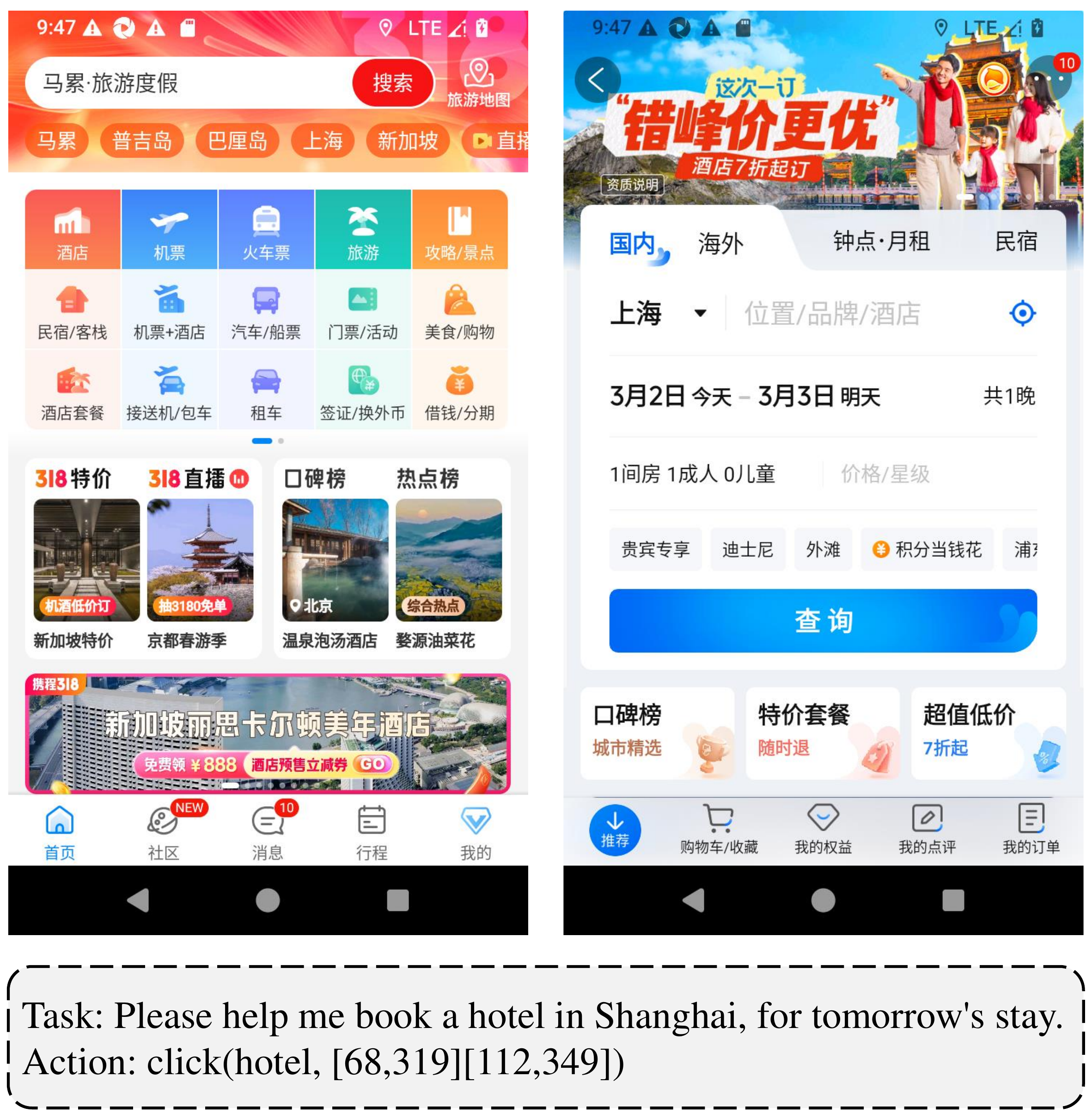}
    \caption{We collect App operation across 50 real world apps.}
    \label{figure1-1}
  \end{minipage}
  \hfill
  \begin{minipage}[b]{0.59\textwidth}
    \centering
    \includegraphics[width=\textwidth]{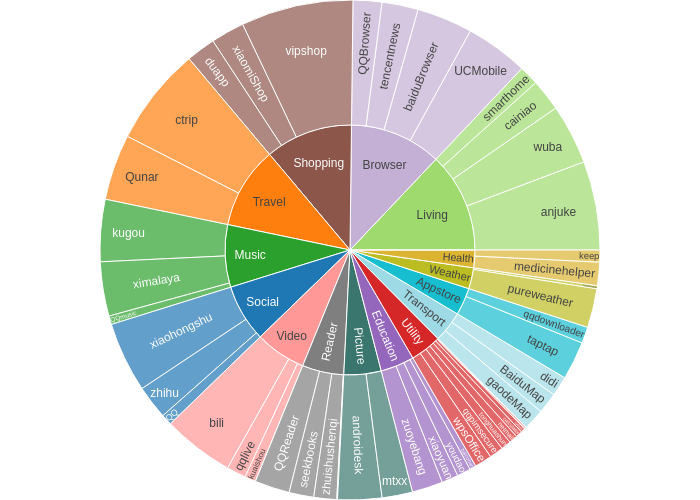}
    \caption{}
    \label{graph}
  \end{minipage}
\end{figure}
\section{Dataset Details}{\label{datadetail}}
\subsection{Background}
\textbf{Environment:} Considering the efficiency and concurrency of data collection, we configured more than 50 simulators on an arm64 architecture server cluster. These emulators are all deployed on the Cuttlefish framework\footnote{\url{https://source.android.google.cn/docs/devices/cuttlefish}}  and are directly accessed and supervised through the web page on Google Cloud Engine. Specifically, each emulator has the same configuration: Android 14 operating system, ARMv8 CPU architecture, 4.75-inch screen with 720x1280 resolution, and 320 DPI. Additionally, during initialization, they are allocated a 6-core CPU and 24GB of memory to ensure the smoothness of random walks on more complex applications.
\begin{figure}[h]
    \centering
    \begin{minipage}[b]{0.48\textwidth}
        \centering
        \includegraphics[width=\textwidth]{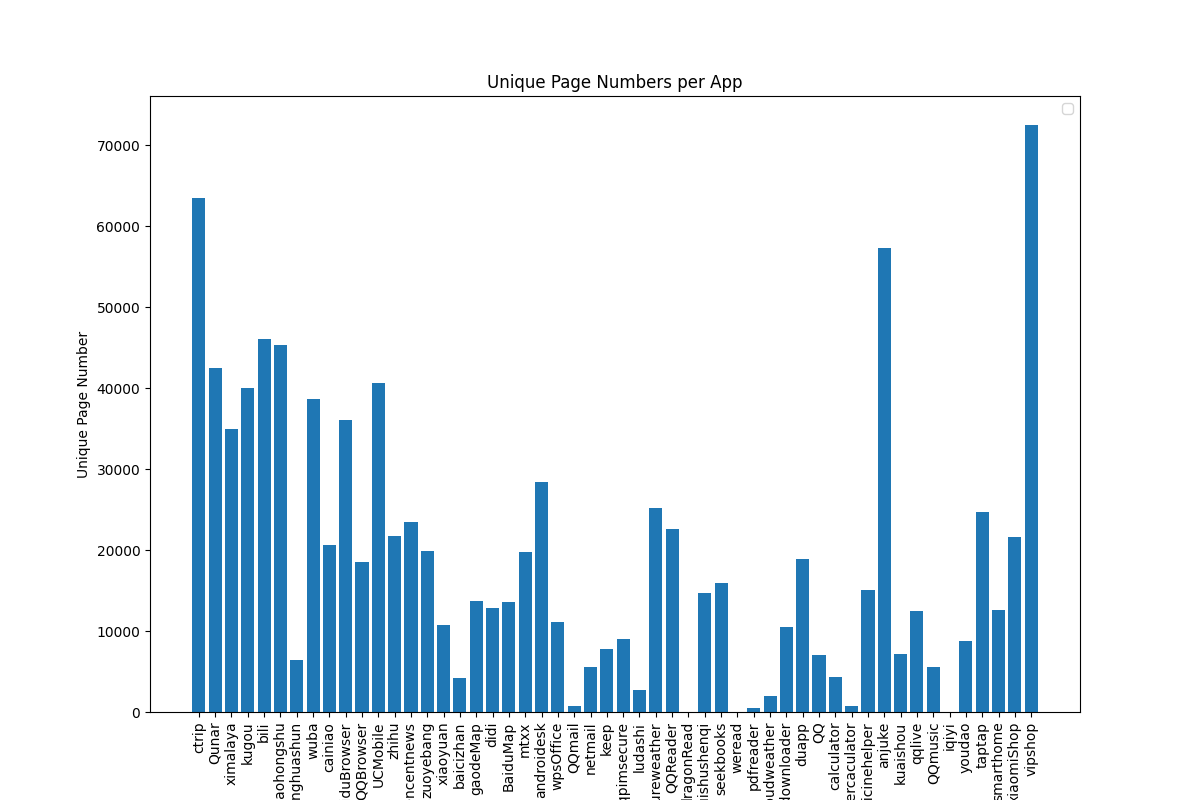}
        \caption{Unique Page Numbers per App}
        \label{fig:unique_page_num}
    \end{minipage}
    \hfill
    \begin{minipage}[b]{0.48\textwidth}
        \centering
        \includegraphics[width=\textwidth]{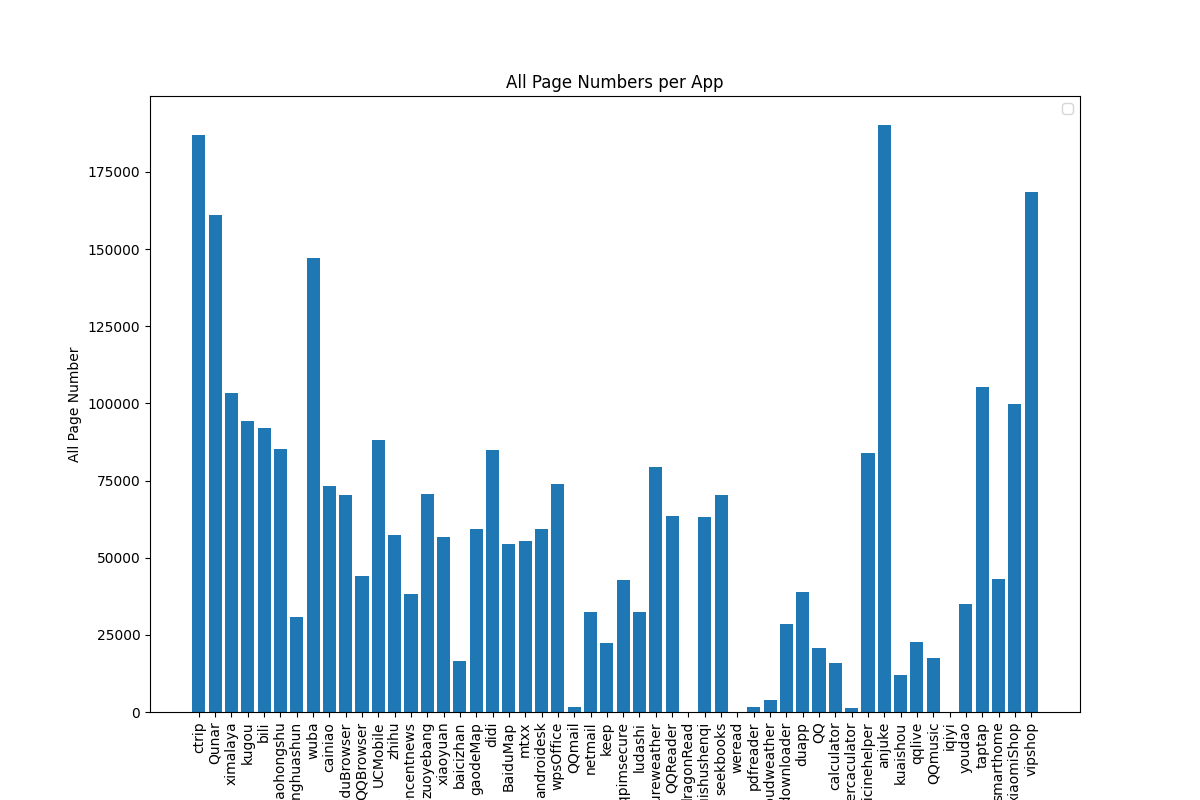}
        \caption{All Page Numbers per App}
        \label{fig:all_page_num}
    \end{minipage}
\end{figure}

\begin{figure}[h]
    \centering
    \includegraphics[width=0.8\textwidth]{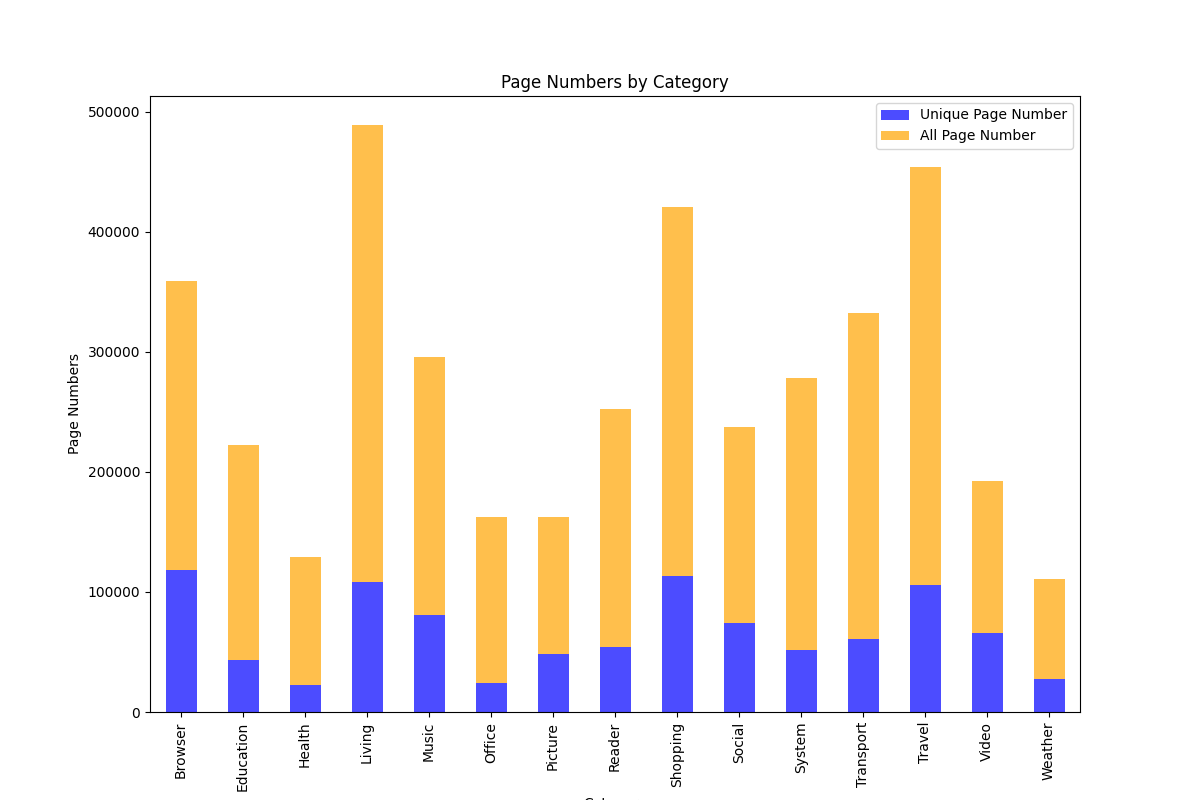}
    \caption{Category Stats}
    \label{fig:category_stats}
\end{figure}
\subsection{MAU of selected APPs}
To demonstrate that the selected apps, although few in number, cover the majority of daily use cases for Chinese users, we can use two key metrics: download volume and Monthly Active Users (MAU). For example, we will use data from the travel and music categories collected from the Tencent App Store.

\begin{table}[h]
\centering
\begin{minipage}{0.45\textwidth}
\centering
\small
\caption{Travel Applications Statistics}
\label{tab:travel}
\begin{tabular}{l|cc}
\hline
\textbf{Application} & \textbf{Downloads (millions)} & \textbf{Ratio} \\ 
\hline
\textbf{Ctrip}       & \textbf{38000}  & \textbf{48.17\%} \\
\textbf{Qunar}       & \textbf{35000}  & \textbf{44.36\%} \\
Tongcheng   & 2600   & 3.30\%  \\
Fantawild   & 165.5  & 0.21\%  \\
Fliggy      & 8008   & 10.15\% \\
Mafengwo    & 2318   & 2.93\%  \\
Aowei       & 95     & 0.12\%  \\
Tuniu       & 8082   & 10.25\% \\
CYTS        & 62.3   & 0.08\%  \\
Disney      & 282.2  & 0.36\%  \\
Spring Tour & 78.6   & 0.10\%  \\
\hline
\end{tabular}
\end{minipage}
\hfill
\begin{minipage}{0.45\textwidth}
\centering
\caption{Music Applications Statistics}
\small
\label{tab:music}
\begin{tabular}{l|cc}
\hline
\textbf{Application} & \textbf{Downloads (millions)} & \textbf{Ratio} \\ 
\hline
\textbf{QQ Music}         & \textbf{274000}  & \textbf{35.24\%} \\
\textbf{Kugou}            & \textbf{292000}  & \textbf{37.56\%} \\
Qishui           & 257.9   & 0.03\%  \\
Kuwo             & 122000  & 15.69\% \\
Tomato Changting & 142.4   & 0.02\%  \\
Huisen           & 43      & 0.01\%  \\
Bodian           & 211.1   & 0.03\%  \\
Migu             & 27000   & 3.47\%  \\
Qianqian Queting & 7872.2  & 1.01\%  \\
Apple Music      & 23.9    & 0.01\%  \\
Free             & 6.6     & 0.01\%  \\
\hline
\end{tabular}
\end{minipage}
\end{table}
From the table \ref{tab:music} and \ref{tab:travel}, we can see that although the number of apps we selected is relatively small, they account for over 70\% of the usage.
Additionally, Mobile3M includes a wide range of app categories that are sufficient to cover users' daily life scenarios. 
The table below shows the number of active users in October 2023 in millions. 
The bolded sections are the apps we have chosen. 
The Monthly Active Users (MAU) data further supports the representativeness of these selected apps. 
\begin{table}[h]
\centering
\begin{minipage}{0.45\textwidth}
\centering
\small 
\caption{Travel Applications Statistics}
\label{tab:travel_new}
\begin{tabular}{l|cc}
\hline
\textbf{Application} & \textbf{MAU (millions)} & \textbf{Ratio} \\ 
\hline
\textbf{Ctrip}              & \textbf{6964.7}  & \textbf{51.96\%} \\
\textbf{Qunar}              & \textbf{2761}    & \textbf{20.60\%} \\
Fliggy             & 1868.3  & 13.94\% \\
Tongcheng          & 574     & 4.28\%  \\
Mafengwo           & 459     & 3.42\%  \\
Huazhu Club        & 351     & 2.62\%  \\
Zhixing            & 120     & 0.90\%  \\
Booking            & 85      & 0.63\%  \\
Ctrip Business     & 116     & 0.87\%  \\
State Grid Business & 104     & 0.78\%  \\
\hline
\end{tabular}
\end{minipage}
\hfill
\begin{minipage}{0.45\textwidth}
\centering
\small 
\caption{Music Applications Statistics}
\label{tab:music_new}
\begin{tabular}{l|cc}
\hline
\textbf{Application} & \textbf{MAU (millions)} & \textbf{Ratio} \\ 
\hline
\textbf{QQ Music}          & \textbf{1869.7}  & \textbf{32.09\%} \\
\textbf{Kugou}             & \textbf{1850.1}  & \textbf{31.75\%} \\
Kuwo              & 929     & 15.94\% \\
Qishui            & 327.1   & 5.61\%  \\
Himalaya          & 680.7   & 11.68\% \\
Qingting          & 60.5    & 1.04\%  \\
Maoer             & 51.9    & 0.89\%  \\
Xiaoyuzhou        & 39      & 0.67\%  \\
Lizhi             & 19.8    & 0.34\%  \\
\hline
\end{tabular}
\end{minipage}
\end{table}

\subsection{Graph Structure}
As shown in Figure \ref{figure3-1}, datasets such as AITW and Rico have extensively collected mobile screen pages, but they have not captured the relationships between these pages. From the perspective of page relationships, they are all point structures. Subsequent works based on these datasets, such as Auto-UI and MoTiF, have constructed action execution traces based on high-level language instructions. These traces are chain structures and maintain basic page relationships. 
However, in practice, all pages of an app should form a graph structure. Many pages are linked from the app's home page and can interconnect, forming a directed cyclic graph. We used breadth-first search combined with pruning and node merging to construct a page graph for each app. Below, we will show the page graph structure of QQ Music in Figure 12.
\begin{figure}
    \centering
    \includegraphics[width=0.8\textwidth]{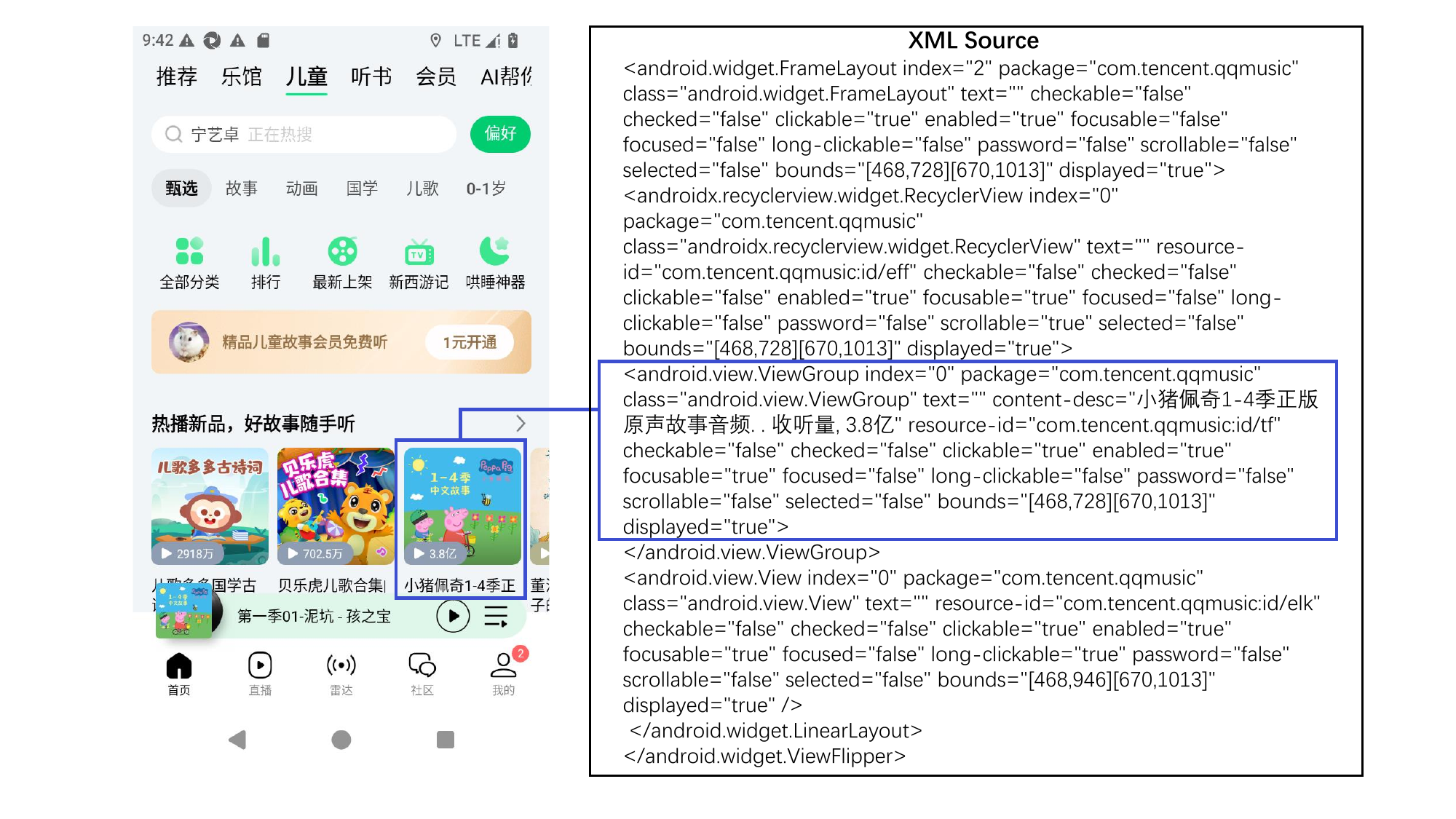}
    \caption{The left part of the figure shows the page, while the right part displays the corresponding XML source file for the image. Due to space constraints, we have excerpted a portion of it.}
    \label{fig_UI_page}
\end{figure}

\subsection{Category Details}
\begin{figure*}[!t]
  \centering
  \includegraphics[width=1\textwidth]{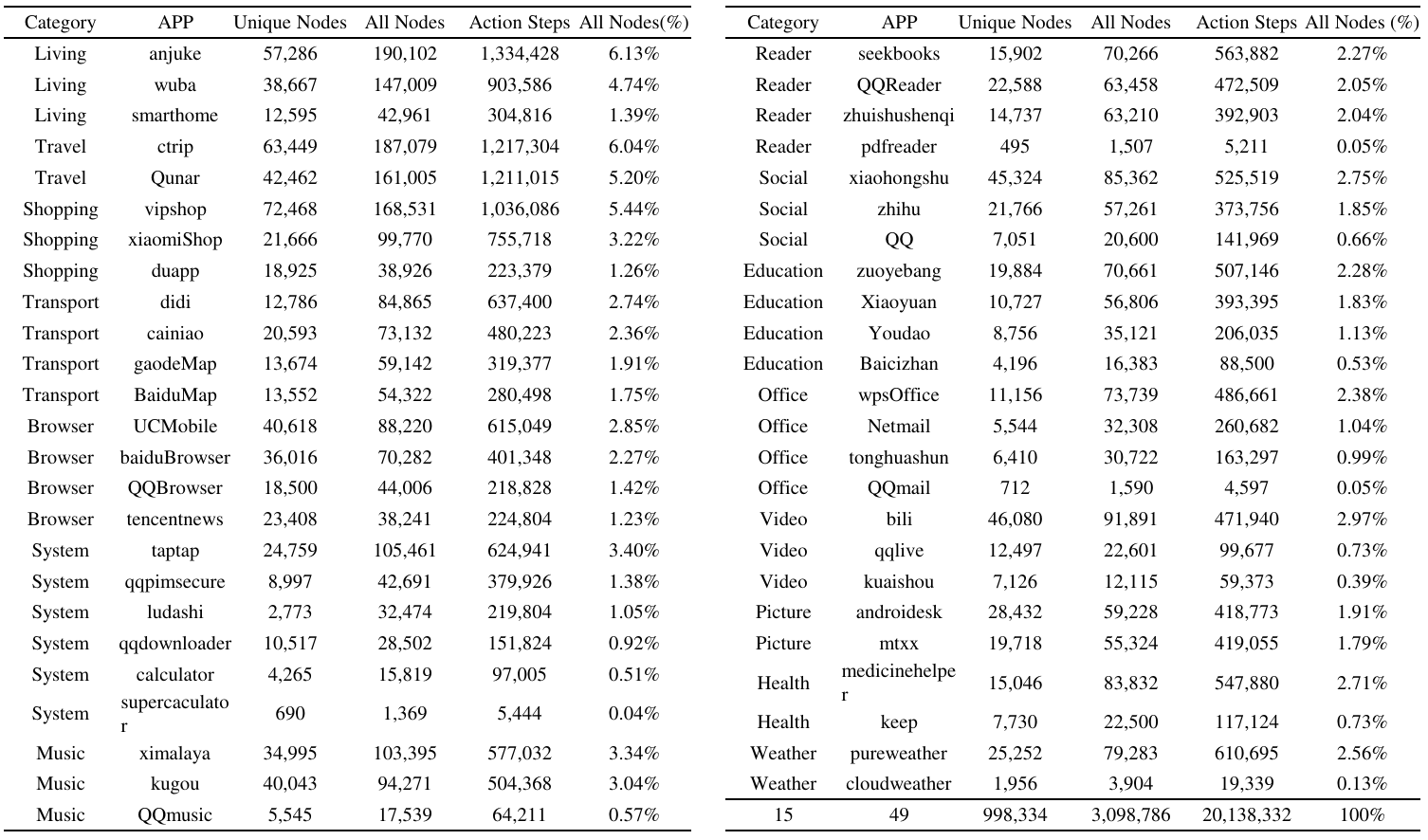}
  \caption{ The data distribution in Mobile3M.}
  \label{dataset_count}
\end{figure*}

Figure \ref{graph} shows a two-layered ring chart. The inner ring represents the data proportion by category, while the outer ring indicates the specific quantity proportion of each app. From this figure, it can be seen that, except for a few apps, most categories have a uniform distribution. For detailed quantity statistics, refer to Figure \ref{dataset_count}.

\section{Experiment results}

\subsection{Stage1-CheckPoints Result}\label{Stage1-CheckPoints}
\begin{figure}[htbp]
    \centering
    \includegraphics[width=0.7\columnwidth]{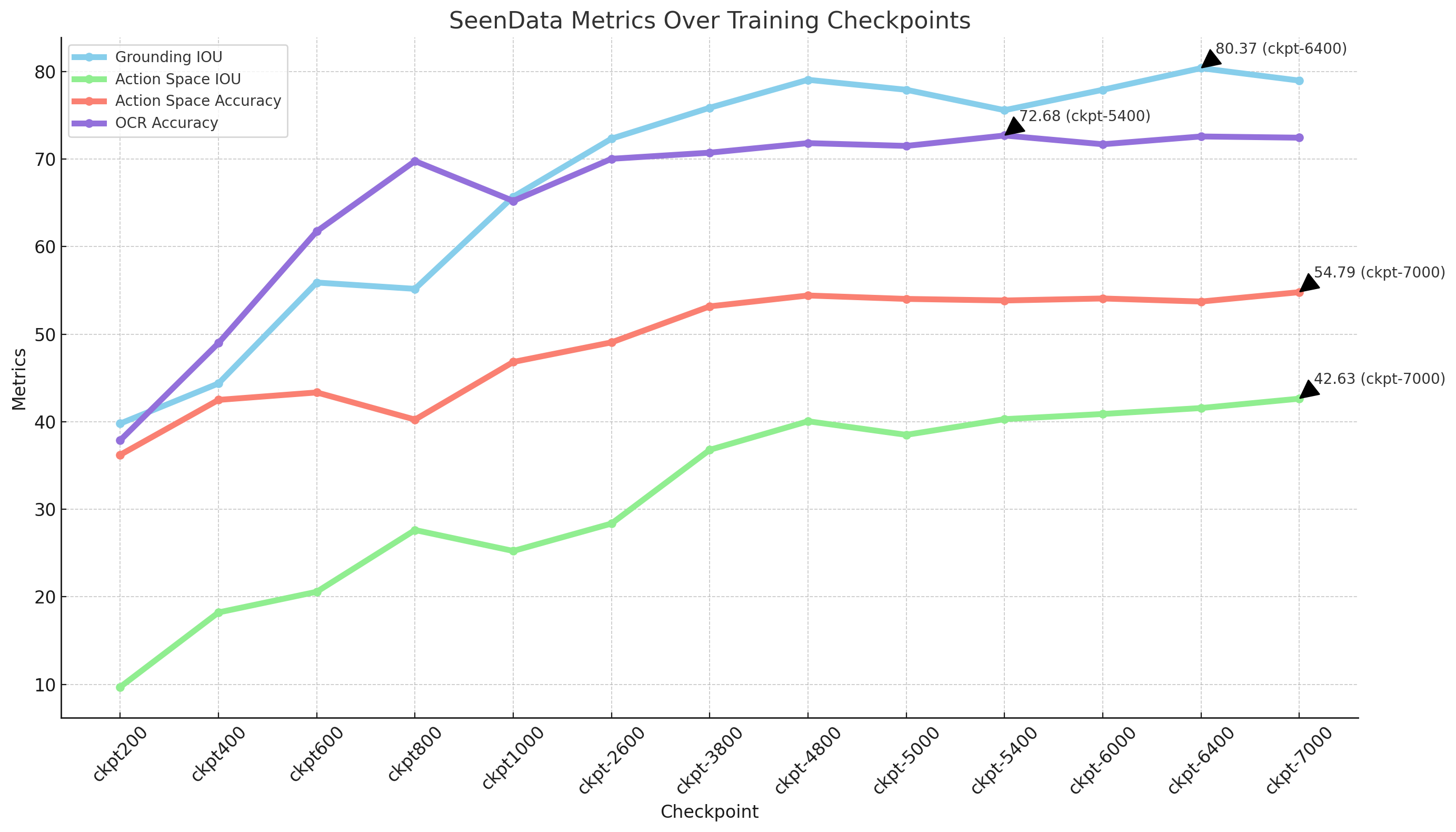}
    \caption{SeenData Metrics Over Training Checkpoints.}
    \label{fig:seen_data_metrics}
\end{figure}

\begin{figure}[htbp]
    \centering
    \includegraphics[width=0.7\columnwidth]{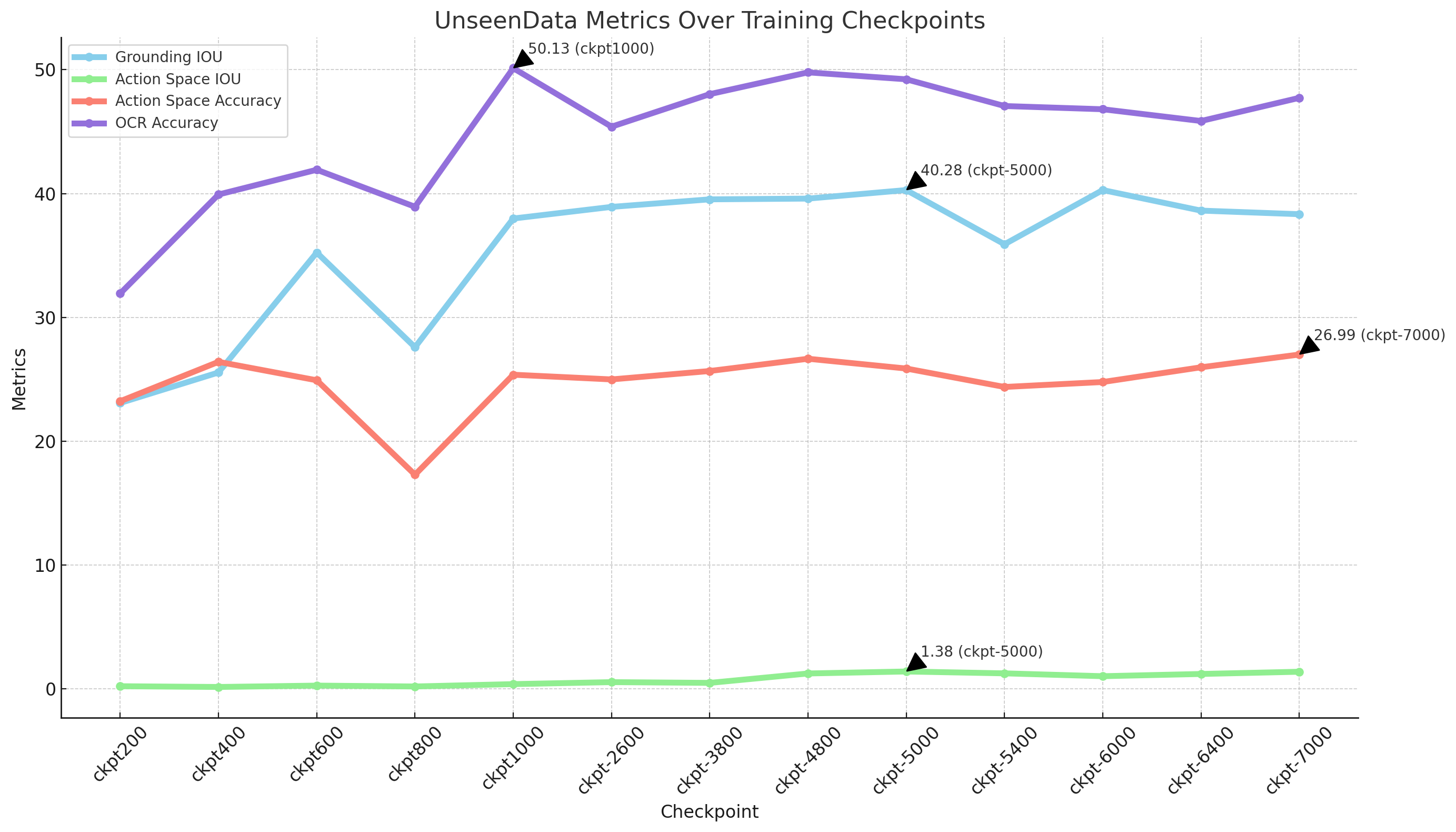}
    \caption{UnseenData Metrics Over Training Checkpoints.}
    \label{fig:unseen_data_metrics}
\end{figure}

Figure \ref{fig:seen_data_metrics} and Figure \ref{fig:unseen_data_metrics} show the test results of the model at different training steps, with the \textbf{black triangle} indicating the best result among the training steps. Due to time constraints, as of writing this paper, we have only completed the current set of training steps, but training is still ongoing. We will continue to update the results in future versions. 
From the figures, it can be observed that the improvement in the ActionSpace Generation task is slow. This is primarily due to the high difficulty of this pre-training task, as the model needs to simultaneously recognize text, determine its location, and identify its widget type. We still consider this task essential because, without additional HTML input, identifying the interaction type of the current widget is a prerequisite for performing the correct action. 
Additionally, upon inspection, the anomalies in the training results were found to be due to uneven distribution of the training data. For detailed data analysis, please refer to Appendix \ref{datadetail}. The abundance of similar types of travel data caused the model to overfit to Ctrip-type apps.


\begin{figure}[htbp]
    \centering
    \includegraphics[width=0.7\columnwidth]{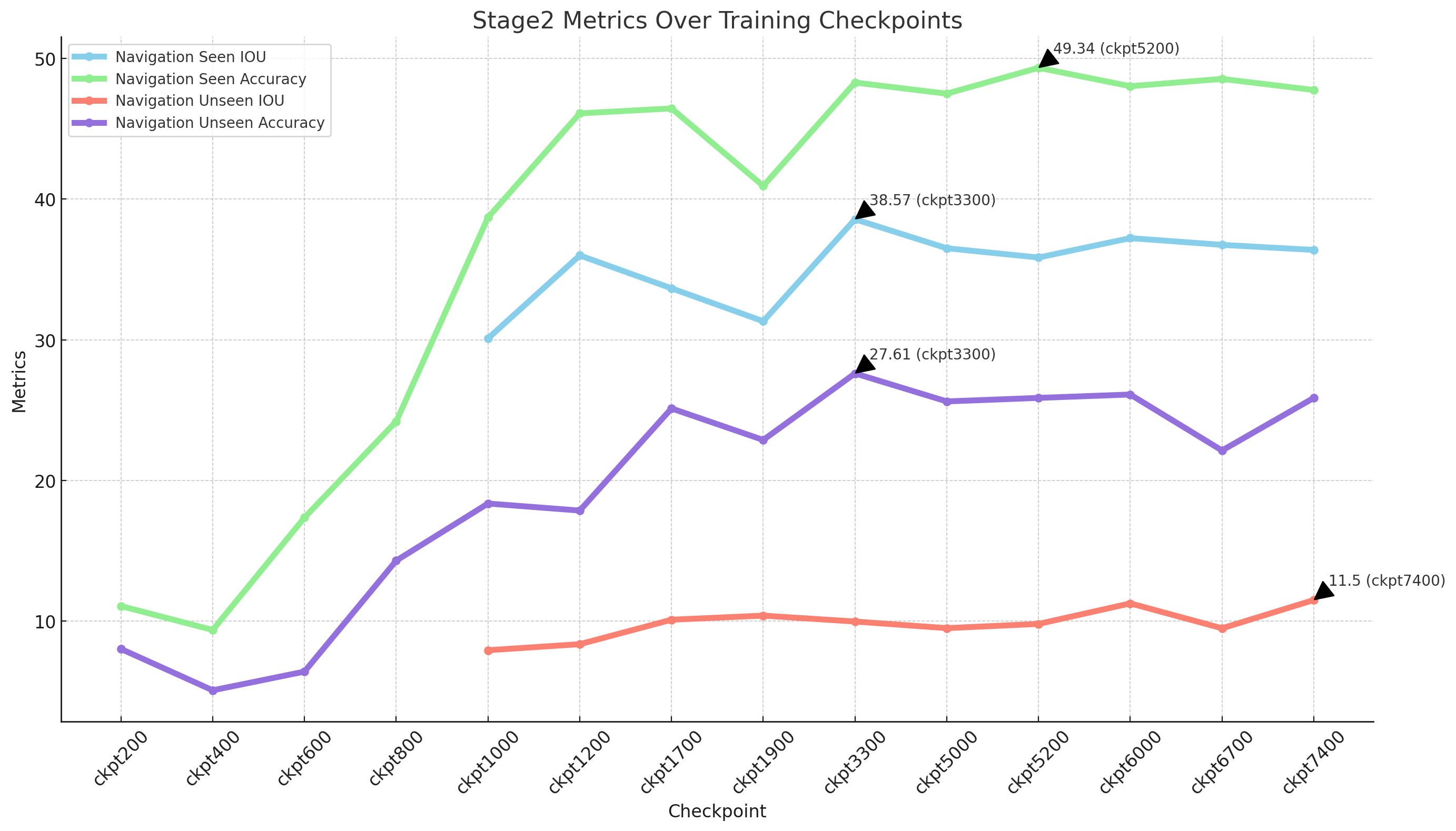}
    \caption{Stage2 Metrics Over Training Checkpoints.}
    \label{fig:stage2_metrics}
\end{figure}

\begin{table*}[!htb]
    \centering
    \resizebox{1\textwidth}{!}{
\begin{tabular}{lcccc|cccc|cc}
\hline
\hline
model(stage1) & Grounding & \multicolumn{2}{c}{Action Space} & OCR & Grounding & \multicolumn{2}{c}{Action Space} & OCR & coco2017 & OCRBench(CN)\\
\hline
          & IoU        & IoU         & Acc         & Acc            & IoU     & IoU         & Acc         & ocr    & IoU & Acc     \\
          \hline
ckpt200 &  39.79    & 9.71 & 36.22 & 37.88 &  23.08 &   0.19  &   23.22  & 31.93 &  -      & -   \\
ckpt400 &   44.37   & 18.22  &  42.49 & 48.98 & 25.54  &     0.13    &  26.40 & 39.92&    -    & -   \\
ckpt600 &  55.89    & 20.58  & 43.34  & 61.74 &  35.23 &     0.24    &  24.91   & 41.92 &     -   & -   \\
ckpt800 & 55.17     & 27.63          & 40.24     & 69.76 & 27.58  & 0.17         & 17.28      & 38.93 &  -  & -   \\  
ckpt1000 & 65.69 & 25.24 & 46.82 & 65.21 & 37.98 & 0.36 & 25.36 & 50.13 & - & - \\  
ckpt-2600 & 72.32  & 28.37 & 49.06 & 70.02 & 38.92 & 0.52 & 24.98 & 45.40  & 22.76 & 19.06 \\  
ckpt-3800 & 75.85  & 36.79 & 53.17 & 70.72 & 39.53  & 0.46 & 25.66 & 48.03 & 18.34 & 24.87 \\  
ckpt-4800 & 79.04  & 40.05 & 54.41 & 71.81 & 39.59  & 1.22 & 26.65 & \text{49.79} & 19.42 & 26.39 \\
ckpt-5000 & 77.90  & 38.50 & 54.02 & 71.49 & 40.28  & \textbf{1.38} & 25.86 & 49.23 & 18.88 & 27.31 \\
ckpt-5400 & 75.57  & 40.29 & 53.84 & \textbf{72.68} & 35.90  & 1.23 & 24.37 & 47.07 & 17.68 & 28.57 \\
ckpt-6000 & 77.90  & 40.88 & 54.07 & 71.68 & 40.28  & 1.0 & 24.77 & 46.81 & 18.64 & 29.86 \\
ckpt-6400 & \textbf{80.37}  & 41.56 & 53.72 & 72.57 & 38.62  & 1.18 & 25.96 &45.86 & 17.19 & 30.04 \\
ckpt-7000 & 78.95  & \textbf{42.63} & \textbf{54.79} & 72.43 & 38.33  & 1.36 & \textbf{26.99} & 47.73 & 17.21 & 30.34 \\
\hline\hline
\end{tabular}
}
    \caption{Stage1 Training Result(\%). }
\end{table*}
\subsection{Stage2-CheckPoints Result}{\label{Stage2-CheckPoints}}

As shown in Figure \ref{fig:stage2_metrics}, the optimal training performance was achieved before 7400 steps, with subsequent training indicating a trend of overfitting. Therefore, we paused the training and selected the CheckPoint3300 version for instruction fine-tuning.
Additionally, from the trend of the test curves, it is evident that the Qwen-VL initially lacks Page Navigation capabilities, but it learns quickly.

\begin{table}[!tb]
    \centering
    \begin{minipage}{0.5\textwidth}
        \centering
        \resizebox{0.9\columnwidth}{!}{
        \begin{tabular}{lcc}
            \hline
            APP Category & Pre-training APP & Test APP\\
            \hline        
            Travel & Ctrip, Amap, Didi & Qunar \\
            Weather & PureWeather & CloudsWeather \\
            Shopping & VIPShop, Xiaomi Mall & DuApp \\
            Reading & QQ Reader & PDF Reader \\
            Email & NetEase Mail & QQ Mail \\
            Dictionary & Youdao & Baicizhan \\
            Books & SeekBooks & Zssq \\
            Music & Kugou Music & QQ Music \\
            Others & Others & - \\
            \hline
        \end{tabular}
        }
        \caption{Pre-training and Test Category}
        \label{test_category}
    \end{minipage}\hfill
    \begin{minipage}{0.5\textwidth}
        \centering
        \resizebox{1\columnwidth}{!}{
        \begin{tabular}{lcc|cc|cc}
            \hline\hline
            model (stage2)           & \multicolumn{2}{c}{Navigation}   & \multicolumn{2}{c}{Navigation} & \multicolumn{2}{c}{MoTIF}  \\
            & IoU     & Acc     & IoU     & IoU     & Acc     & IoU                \\ \hline
            ckpt200 & --- & 11.07 & --- & 8.02 & --- \\
            ckpt400 & --- & 9.37 & --- & 5.08 & --- \\
            ckpt600 & --- & 17.36 & --- & 6.41 & --- \\
            ckpt800 & --- & 24.2 & --- & 14.3 & --- \\
            ckpt1000 & 30.1 & 38.7 & 7.93 & 18.36 & ---\\
            ckpt1200 & 36 & 46.1 & 8.36 & 17.86 & ---\\
            ckpt1700 & 33.66 & 46.45  & 10.10 & 25.12 & --- \\ 
            ckpt1900 & 31.32 & 40.94 & 10.39 & 22.88 & --- \\
            ckpt3300 & \textbf{38.57} & 48.29 & 9.97 & \textbf{27.61} & --- \\ 
            ckpt5000 & 36.51 & 47.50  & 9.50 & 25.62 & --- \\ 
            ckpt5200 & 35.85 & \textbf{49.34} & 9.80 & 25.87 & --- \\ 
            ckpt6000 & 37.23 & 48.03 & 11.26 & 26.11 & ---  \\ 
            ckpt6700 & 36.75 & 48.55  & 9.49 & 22.13 & --- \\
            ckpt7400 & 36.39 & 47.76 & \textbf{11.50} & 25.87  & ---\\
            \hline
        \end{tabular}
        }
        \caption{Stage2 Training Result(\%). }
    \end{minipage}
\end{table}

\section{Case Study}\label{casestudy}
In this section, we discuss several examples that demonstrate how the inclusion of Mobile3M data, without mixing in the original training corpus during the training phase, can affect the model's performance on general tasks. 
\subsection{Illusrtration for Training Consistency}
The consistency here mainly includes two aspects:
(1) Both our pre-training data and fine-tuning domain data are sourced from the mobile domain.
(2) The tasks during both the pre-training and fine-tuning stages are specifically designed for the mobile domain. Below, we will use examples to illustrate this issue. 
\begin{table}[h]
\centering
\caption{Comparison of Training Frameworks}
\label{tab:training_framework}
\begin{tabular}{l|p{0.35\textwidth}p{0.35\textwidth}}
\hline
\textbf{} & \textbf{Previous Training Framework} & \textbf{Mobile3M Training Framework} \\ 
\hline
\textbf{Pre-training stage1} & Grounding tasks on refCOCO. & Element Grounding tasks on Mobile3M. \\
\textbf{Data Example} &``Where is the girl petting the dog?'' & ``Where is the input control containing Hangzhou in the screenshot?'' \\
\hline
\textbf{Pre-training stage2} & VQA tasks on ChartQA. & Action Prediction tasks on Mobile3M. \\
\textbf{Data Example} & ``What is the result of the addition calculation in the table?'' & ``What action should I take to go from image one to image two?'' \\
\hline
\textbf{Fine-tuning} & Instruction navigation tasks on Auto-UI. & Instruction navigation tasks on Mobile3M. \\
\textbf{Data Example} & ``How should I open the alarm clock?'' & ``How should I open the alarm clock?'' \\
\hline
\end{tabular}
\label{training framework}
\end{table}

\subsection{Grounding Task} 
\begin{figure*}[htbp]
    \centering
    \begin{minipage}[t]{0.24\textwidth}
        \centering
        \includegraphics[width=\textwidth]{./picture/checkpoint800_qwen.png}
        \caption{Qwen-VL' result on COCO Grounding Task.}
        \label{fig:qwencoco}
    \end{minipage}
    \hfill
    \begin{minipage}[t]{0.24\textwidth}
        \centering
        \includegraphics[width=\textwidth]{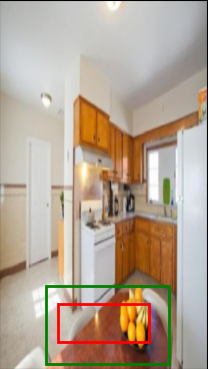}
        \caption{Stage1 Pretraining model's result on COCO Grounding Task.}
        \label{fig:ourscoco}
    \end{minipage}
    \begin{minipage}[t]{0.24\textwidth}
        \centering
        \includegraphics[width=\textwidth]{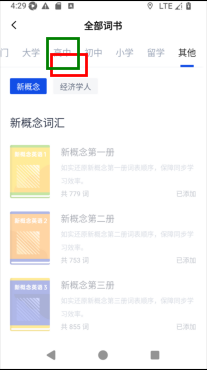}
        \caption{Qwen-VL' result on Mobile3M Grounding Task.}
        \label{fig:qwenmobile}
    \end{minipage}
    \hfill
    \begin{minipage}[t]{0.24\textwidth}
        \centering
        \includegraphics[width=\textwidth]{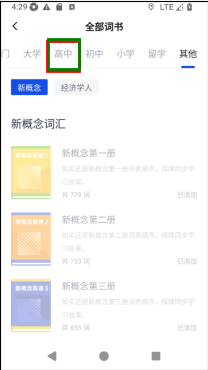}
        \caption{Stage1 Pretraining model's result on Mobile3M Grounding Task.}
        \label{fig:oursmobile}
    \end{minipage}
\end{figure*}

Comparing Figure \ref{fig:qwencoco} and Figure \ref{fig:ourscoco}, it can be observed that Qwen-VL performs well on the original object detection tasks. However, after undergoing Stage 1 training, the LLM experienced a decline in performance on these tasks. 
However, comparing Figures \ref{fig:oursmobile} and \ref{fig:oursmobile}, it can be seen that the model trained with Stage 1 shows improvement on the Mobile3M Grounding task compared to the original Qwen-VL. In the selected examples, Qwen had significant errors, but Stage 1 provided a corrective effect across a broader range of examples. Considering that the data is suitable for training specialized models in the Mobile domain, the performance loss in the general domain is acceptable.
\subsection{Pre-training Task Test Case}
\textbf{1. Language Command Grounding} Qwen-VL-Max and Qwen-VL-Plus have an input limit of 6k tokens. After encoding, a 720×1280 image occupies approximately 2100+ tokens. This means that in most tests, the number of few-shot examples is limited to a single image, resulting in highly unstable outputs. GPT-4o does not have this issue, as it supports a maximum input of 32k tokens, allowing up to 10 images to be processed simultaneously.

\begin{figure*}[!t]
  \centering
  \begin{minipage}[t]{0.50\textwidth}
    \centering
    \includegraphics[width=\textwidth]{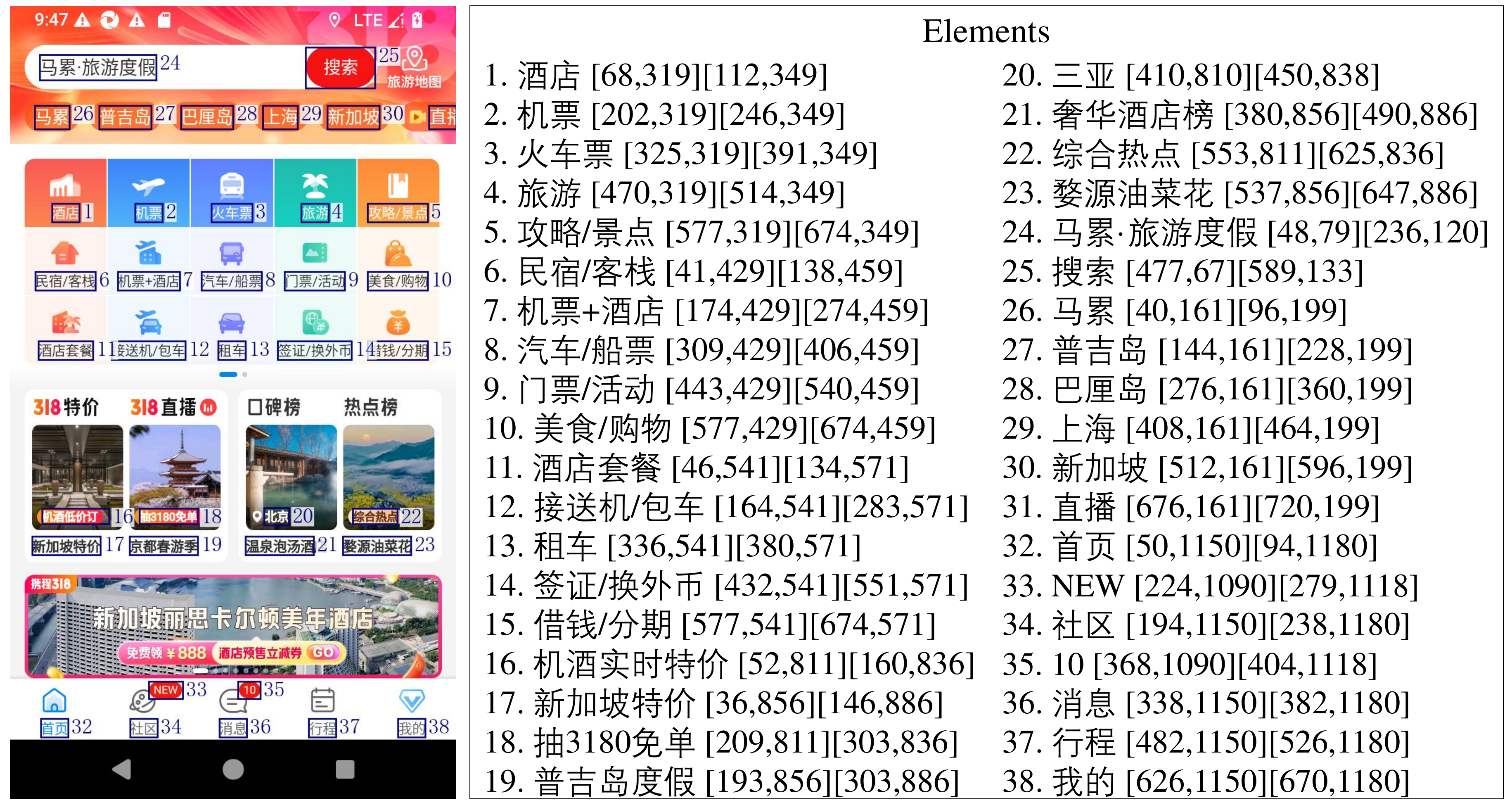}
    \caption{Element List.}
    \label{figure2-1}
  \end{minipage}
  \hfill
  \begin{minipage}[t]{0.46\textwidth}
    \centering
    \includegraphics[width=\textwidth]{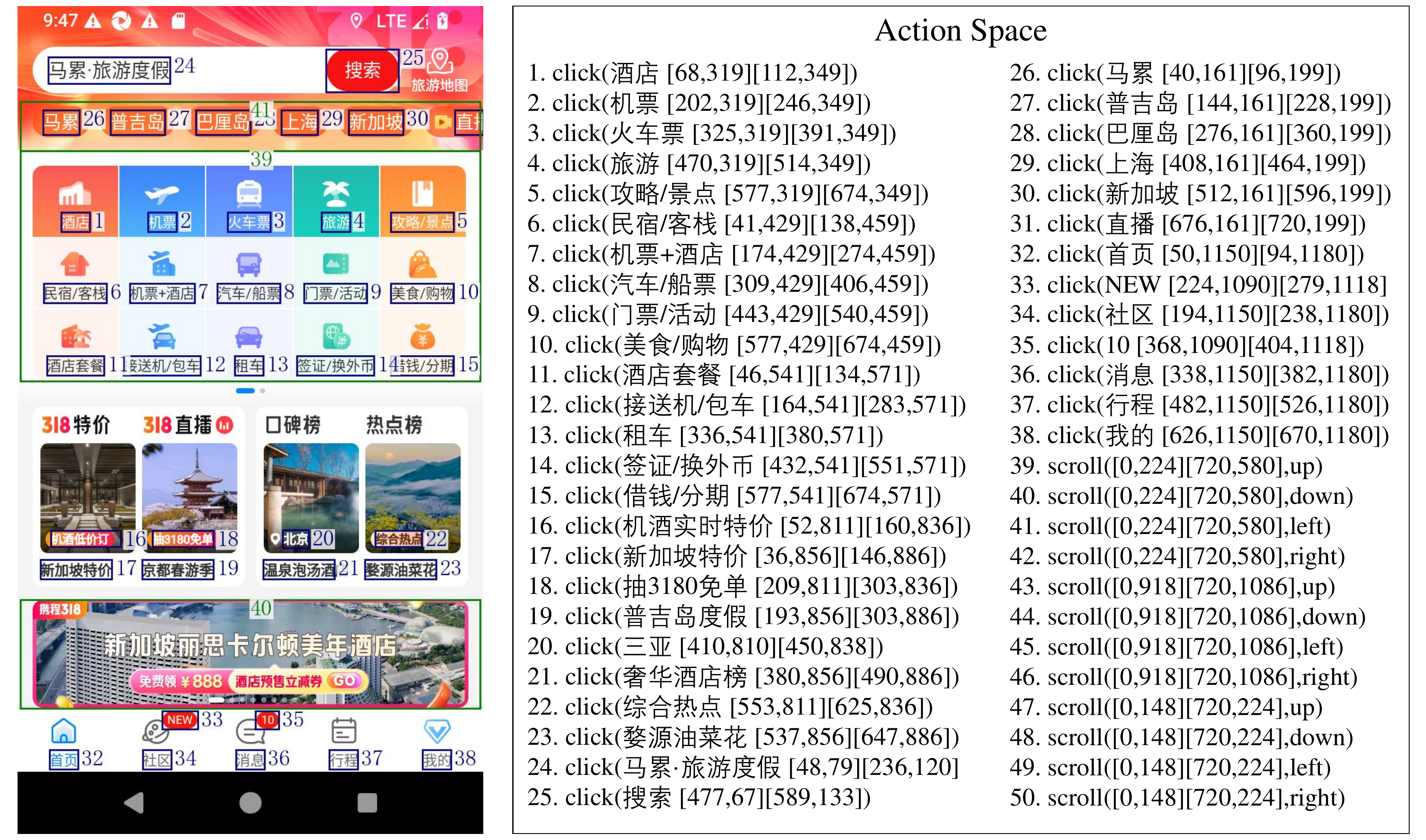}
    \caption{Samples of pre-train data for Action Space Extraction.}
    \label{figure2-2}
  \end{minipage}
\end{figure*}

\textbf{2. Element List Generation}  
As shown in Figure \ref{figure2-1}, in this task, we aim for the model to recognize all the text on the current page, as this is the foundation for interacting with these texts. It is important to note that not all text is interactive, as some may simply be TextView elements or text within images (where the image itself is not clickable).

\textbf{3. Action Space Generation} 
As shown in Figure \ref{figure2-2}, based on the Element List, we expect the model to further distinguish the type of each widget. Since scroll elements are not visible themselves, we manually generated four alternative actions for them: up, down, left, and right. Our action space does not support diagonal scrolling, as this often implies a drag action in practical operations.
Therefore, the model actually only needs to distinguish between input fields and buttons. We also expect it to ground these buttons accurately.
\subsection{Fine-tuning Task Case Study}
The limitation of the MobileVLM-unified model in the Auto-UI single task: Multi-step mixed training tasks cause the model to make more errors in determining the end of a task.
Below are the different responses from MobileVLM-unified and MobileVLM-separate when completing the same task: 




\end{document}